\documentclass[11pt, a4paper, logo, copyright]{googlecloud}

\pdftrailerid{redacted}

\makeatletter
\renewcommand\bibentry[1]{\nocitep{#1}{\frenchspacing\@nameuse{BR@r@#1\@extra@b@citeb}}}
\makeatother

\usepackage[authoryear, sort&compress, round]{natbib}

\usepackage{latexsym}

\usepackage[T1]{fontenc}
\usepackage[utf8]{inputenc}

\usepackage{microtype}

\usepackage{graphicx}

\usepackage[utf8]{inputenc} % allow utf-8 input
\usepackage[T1]{fontenc}    % use 8-bit T1 fonts
\usepackage{hyperref}       % hyperlinks
\usepackage{url}            % simple URL typesetting
\usepackage{booktabs}       % professional-quality tables
\usepackage{amsfonts}       % blackboard math symbols
\usepackage{nicefrac}       % compact symbols for 1/2, etc.
\usepackage{microtype}      % microtypography
\usepackage{xcolor}         % colors
\usepackage{graphicx}
\usepackage{algorithm}
\usepackage{algorithmic}
\usepackage{amsmath}
\usepackage{multicol}
\usepackage{multirow}
\usepackage{enumitem}
\usepackage{float}
\usepackage{tcolorbox}
\tcbuselibrary{breakable}
\usepackage{tcolorbox}
\tcbuselibrary{breakable} % Required to allow page breaks
\usepackage{graphicx}
\usepackage{xcolor}
\usepackage{wrapfig}
\usepackage{titletoc}
\usepackage{colortbl}
\usepackage{graphicx}
\usepackage{algorithm}
\usepackage{algorithmic}
\usepackage{amsmath}
\usepackage{multicol}
\usepackage{multirow}
\usepackage{enumitem}
\usepackage{float}
\usepackage{tcolorbox}
\tcbuselibrary{breakable}
\usepackage{graphicx}
\usepackage{wrapfig}
\usepackage{subcaption}
\usepackage{kantlipsum, lipsum}
\usepackage{dsfont}

\usepackage[utf8]{inputenc} % allow utf-8 input
\usepackage[T1]{fontenc}    % use 8-bit T1 fonts
\usepackage{hyperref}       % hyperlinks
\usepackage{url}            % simple URL typesetting
\usepackage{booktabs}       % professional-quality tables
\usepackage{amsfonts}       % blackboard math symbols
\usepackage{nicefrac}       % compact symbols for 1/2, etc.
\usepackage{microtype}      % microtypography
\usepackage{xcolor}         % colors
\usepackage{graphicx}
\usepackage{algorithm}
\usepackage{algorithmic}
\usepackage{amsmath}
\usepackage{multicol}
\usepackage{multirow}
\usepackage{enumitem}
\usepackage{float}
\usepackage{tcolorbox}
\tcbuselibrary{breakable}
\usepackage{tcolorbox}
\tcbuselibrary{breakable} % Required to allow page breaks
\usepackage{graphicx}
\usepackage{xcolor}
\usepackage{wrapfig}
\usepackage{titletoc}

\newcommand{\ours}{\textsc{STRIDE}}

\definecolor{added}{rgb}{0,0.6,0}
\definecolor{deleted}{rgb}{1,0,0}

\newcommand{\hmap}[2]{%
    \pgfmathsetmacro{\percent}{((#1 - 1) / 4) * 100}%
    \ifdim \percent pt > 50 pt
        \def\mytextcol{white}%
    \else
        \def\mytextcol{black}%
    \fi
    \edef\temp{\noexpand\cellcolor{green!\percent!white}}\temp%
    \textcolor{\mytextcol}{$#1_{#2}$}%
}
\theoremstyle{plain}
\newtheorem{theorem}{Theorem}[section]
\newtheorem{proposition}[theorem]{Proposition}

\theoremstyle{definition}

\theoremstyle{remark}

\newcommand{\benchmark}[1]{\textit{TFRBench}}
\tcbuselibrary{breakable}

\definecolor{verylightgray}{gray}{0.9}

\definecolor{light_red}{rgb}{1.0, 0.6, 0.6}
\definecolor{light_green}{rgb}{0.56, 0.93, 0.56}

\title{STRIDE: Reasoning-Aware Training for Time Series Forecasting}

\correspondingauthor{Mihir Parmar \href{mailto:mihirparmar@google.com}{<mihirparmar@google.com>}, Md Atik Ahamed \href{mailto:atikahamed@google.com}{<atikahamed@google.com>}, and Palash Goyal \href{palashgoyal@google.com}{<palashgoyal@google.com>}}

\renewcommand{\today}{}

\author[1, 2]{\fontsize{10.0pt}{10.0pt}\selectfont Md Atik Ahamed}
\author[1]{\fontsize{10.0pt}{10.0pt}\selectfont Mihir Parmar}
\author[1]{\fontsize{10.0pt}{10.0pt}\selectfont Palash Goyal}
\author[1]{\fontsize{10.0pt}{10.0pt}\selectfont Chun-Liang Li}
\author[2]{\fontsize{10.0pt}{10.0pt}\selectfont Qiang Cheng}
\author[1]{\fontsize{10.0pt}{10.0pt}\selectfont Tomas Pfister}
\author[1]{\fontsize{10.0pt}{10.0pt}\selectfont Jinsung Yoon}
\affil[1]{\fontsize{9.0pt}{9.0pt}\selectfont Google}
\affil[2]{\fontsize{9.0pt}{9.0pt}\selectfont University of Kentucky}

\begin{document}
\begin{abstract}

Time Series Foundation Models (TSFMs) excel at numerical forecasting but operate as black boxes lacking qualitative reasoning. Conversely, applying LLMs directly to temporal data introduces a modality gap: text tokenizers fragment continuous numerical values, degrading mathematical relationships and exploding sequence lengths, leading to computational overhead. To resolve this, we introduce \ours{} (Strategic Time-series Reasoning Injected via Distilled Embeddings), a novel framework natively integrating LLM reasoning into the continuous embedding space of TSFMs. Instead of discrete tokens, \ours{} distills reasoning traces into a lightweight LLM and dynamically projects its pooled hidden states as a continuous cross-modal prior into the numerical encoder. Jointly optimized via cross-entropy and quantile losses, \ours{} readily attaches to diverse architectures (e.g., Chronos-2, Timer-S1). \ours{} ranks \textbf{\#1 on the GIFT-Eval leaderboard (0.625 MASE, 0.423 CRPS)} when ensembled with Synapse. On TFRBench, \ours{} consistently outperforms baselines in both forecasting accuracy and reasoning quality, with ablations confirming gains stem from strategic reasoning. Thus, injecting semantic reasoning as a continuous prior equips TSFMs with human-interpretable reasoning while improving predictive accuracy.

\end{abstract}

\maketitle

\section{Introduction}

Time Series Foundation Models (TSFMs) have recently transformed numerical forecasting by applying large-scale pretraining to temporal data, establishing new baselines in numerical forecasting \citep{ansari2024chronos, ansari2025chronos, das2024decoder, liu2026timer, woo2024unified}. However, their success relies entirely on acting as black-box mathematical extrapolators. While they effectively capture historical dependencies (e.g., trends), they fundamentally lack the ability to process semantic context, articulate predictive reasoning, or dynamically adapt to external variables such as holidays or economic shifts. Since TSFMs approximate the conditional distribution $P(Y|X)$ using only continuous numerical representations, they suffer from a blind spot in qualitative reasoning.

Motivated by this limitation, recent literature has attempted to directly apply LLMs to time series forecasting \citep{tang2025time, ye2024mirai}. Yet, emerging studies reveal a profound ``LLM mirage''--using only LLMs for forecasting often fails to meaningfully outperform TSFMs \citep{tan2024are}. Prior work attributes this in part to a modality gap, where forcing continuous temporal signals through discrete text tokenizers can fragment numerical values, degrading mathematical precision while introducing computational bottlenecks \citep{11435839, tao2025values}. LLMs are fundamentally designed for semantic reasoning, not for autoregressive numerical calculation.

\begin{figure*}
    \centering
    \includegraphics[width=\linewidth,trim=0mm 1mm 3mm 0mm, clip]{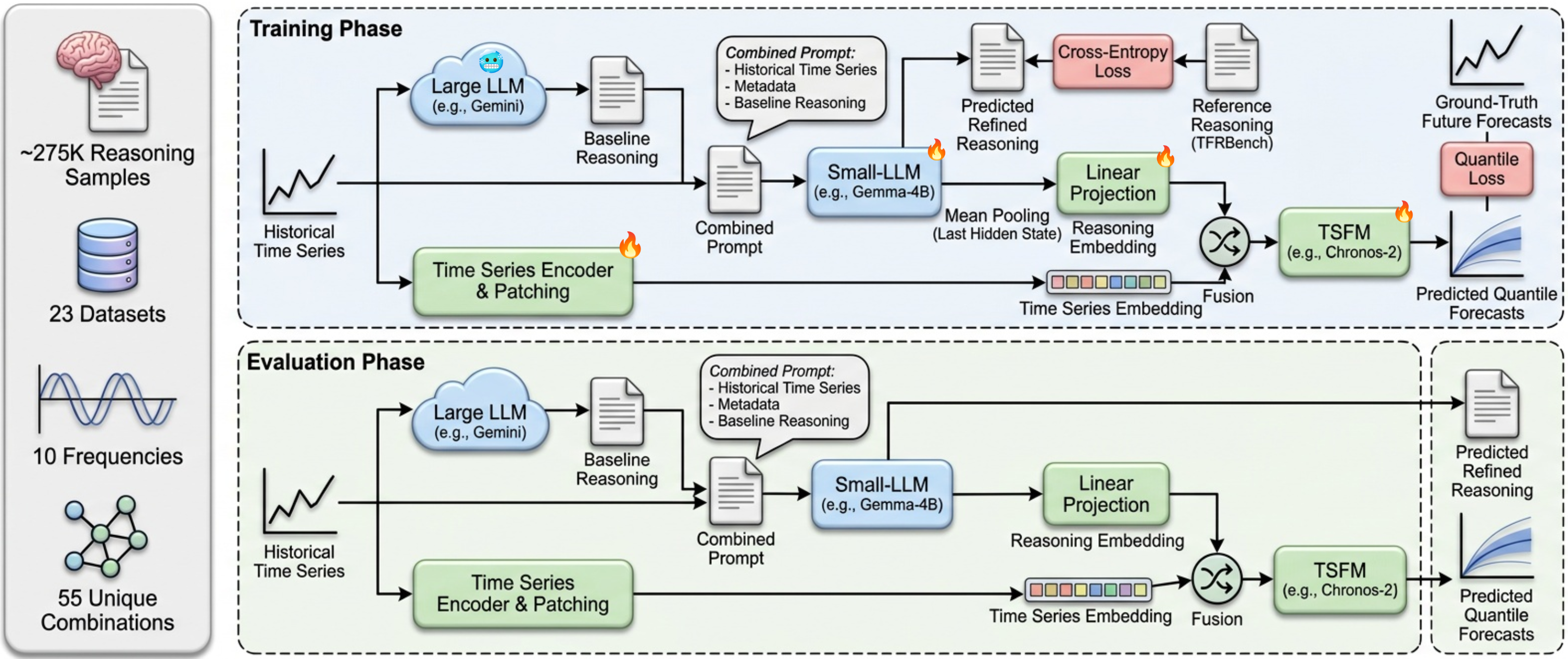}
    \caption{\textbf{Overview of the \ours{}. (Left)} Diverse training corpus of $\sim275K$ reasoning samples across 23 datasets. \textbf{(Right)} The pipeline is jointly optimized via cross-entropy and quantile losses during training, enabling the framework to simultaneously output human-interpretable reasoning and numerical forecasts during evaluation. \protect\includegraphics[height=1em]{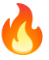} and \protect\includegraphics[height=1em]{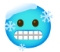} indicate trainable and frozen modules, respectively.}
    \label{fig:teaser}
    % \vspace{-.3cm}
\end{figure*}

In this work, we argue that LLMs should not be forced to act as numerical forecasters; instead, their utility lies in generating statistically grounded meaningful reasoning. To this end, we reformulate the forecasting objective to model the conditional distribution $P(Y|X,R)$, where $R$ acts as an explicit reasoning prior (motivated from \citet{ahamed2026tfrbench}). To formalize this new objective, we introduce \ours{} (illustrated in Figure \ref{fig:teaser}), a novel framework that fuses semantic reasoning priors from LLMs with continuous numerical prediction by integrating it directly into TSFMs.

\vspace{0.09cm}

\ours{} relies on cross-modal representation distillation \citep{hsieh-etal-2023-distilling, tian2025beyond}. During the training phase, a teacher model with access to reference ground-truth data and external events generates oracle reasoning, motivated by \citet{ahamed2026tfrbench}. We distill this reasoning generation capability into a lightweight student LLM, teaching it to articulate equivalent insights using strictly historical context and time series. Oracle access to future values and retrieved events is used exclusively to construct training targets, preventing any temporal leakage into the system at test time. Crucially, to bypass the token-limit bottlenecks and information loss of previous works \citep{tao2025values}, we avoid passing discrete text tokens to the forecasting model. Instead, we extract the continuous mean-pooled hidden states of the student LLM and project them directly into the target TSFM's embedding space. The pipeline is jointly optimized end-to-end via textual cross-entropy \citep{brown2020language} and numerical quantile \citep{ansari2024chronos, lim2021time} losses.

We conduct evaluations across two large-scale benchmarks to validate both the numerical and reasoning performance of \ours{}. First, on the GIFT-Eval \citep{aksu2025gifteval}, \ours{} achieves 0.674 MASE and 0.454 CRPS, outperforming the compared TSFMs (Figure~\ref{fig:gifteval_result}). Second, to assess generalizability and reasoning quality, we evaluate on TFRBench \citep{ahamed2026tfrbench}, where \ours{} demonstrates superior numerical performance against the baselines in both in-domain (0.615 vs. 0.765 MASE) and out-of-domain (0.724 vs. 0.778 MASE) settings on the majority of the scenarios. Beyond numerical performance, an LLM-as-a-Judge evaluation confirms that \ours{} achieves the highest scores on the majority of cases of the four qualitative metrics, consistently generating logic-grounded reasoning that outperforms zero-shot LLM forecasters. Ablation studies reveal that \ours{} is model agnostic: it can be attached to diverse TSFMs (e.g., Chronos-2.0, Timer-S1) and consistently improves their performance. Further ablation experiments (\textsection \ref{sec:ablation}, and App. \ref{app:additional_ablations}) show that numerical fine-tuning without reasoning, shuffled series--reasoning pairing, generic summary embeddings, and text-to-number TS-LLMs (e.g., PromptCast) remain higher-error than \ours{}. Thus, \ours{} points toward more interpretable foundation models for forecasting.

In summary, our contributions are threefold: (i) a \textbf{Novel Cross-Modal Framework} that utilizes a generalized latent projection mechanism to natively inject LLM reasoning into TSFM embedding spaces, bypassing discrete token bottlenecks; (ii) a \textbf{Reasoning Distillation Pipeline} enabling a lightweight student LLM to generate high-quality analytical strategies using strictly historical context; and (iii) \textbf{Better Numerical and Reasoning Performance}, empirically demonstrating improved performance on GIFT-Eval, and TFRBench compared to various baselines with detailed ablations.

\section{Related Works}

\paragraph{TSFMs.}
The success of the Transformer architecture in NLP has spurred the development of TSFMs designed for zero-shot generalization. Models such as TimesFM \citep{das2024decoder}, Chronos \citep{ansari2024chronos,ansari2025chronos}, MOIRAI \citep{woo2024unified}, Timer-S1 \citep{liu2026timer} and Lag-Llama \citep{rasul2024lagllama} have achieved state-of-the-art performance. These approaches typically employ specialized tokenization strategies, either patching numerical values \citep{nie2023a} or discretizing them into vocabulary tokens \citep{gruver2024llmtime} to process time series as a language. While effective at capturing complex autoregressive dependencies and scaling laws, they do not natively leverage external semantic knowledge (e.g., holiday impacts or economic shifts) or qualitative reasoning capabilities. Conversely, recent attempts to replace TSFMs entirely with general-purpose LLMs have exposed a profound modality gap \citep{gruver2024llmtime}, where discrete text tokenization degrades mathematical precision. Our work bridges this divide: rather than replacing TSFMs, we introduce a joint optimization framework that trains both the numerical forecaster and the language model, utilizing a latent projection mechanism to inject the structural reasoning of the LLM directly into the continuous embedding space of the TSFM.

\paragraph{Prompt-Driven and Text-to-Text Forecasting.}
A growing body of work explores using general-purpose LLMs (e.g., Gemini, Claude) for forecasting by prompting them to analyze data. PromptCast \citep{xue2023promptcast} was among the first to frame forecasting as a sentence-to-sentence generation task, and subsequent works in time-series have introduced Chain-of-Thought (CoT)~\citep{chow2024towards} or programmatic ways~\citep{jin2024timellm} to elicit intermediate analytical steps. However, relying on discrete text tokens for numerical prediction introduces computational bottlenecks and information loss. Furthermore, existing CoT forecasting research suffers from a lack of rigorous evaluation; LLMs often ``hallucinate'' causal links to justify arbitrary predictions \citep{ji2023survey}, and traditional benchmarks fail to penalize this disconnect. STRIDE addresses this by distilling high-quality, verified reasoning from a teacher model into a lightweight student model, jointly optimizing textual generation and numerical forecasting so that numerical decoding remains in the TSFM rather than in text tokens.

Extended related work explaining dataset-specific architectures, explainable forecasting are provided in App.~\ref{app:extend_related}.

\section{\ours{}}
\label{sec:method}

Figure~\ref{fig:teaser} and Algorithm \ref{alg:method} illustrate the overall framework, which consists of a textual reasoning branch and a numerical forecasting branch fused via a generalized latent projection mechanism.

\subsection{Problem Formulation and Motivation}
Given a historical multivariate time series context $\mathbf{X} \in \mathbb{R}^{T \times V}$, where $T$ is the context length and $V$ is the number of variates, our objective is to predict the future trajectory $\mathbf{Y} \in \mathbb{R}^{H \times V}$ for a forecasting horizon $H$. Standard numerical models approximate the conditional distribution $P(\mathbf{Y} \mid \mathbf{X})$. To enhance both performance and interpretability, we introduce an explicit reasoning prior $\mathbf{R}$, capturing cross-channel dynamics, trend shifts, and periodic events in a structured format. Consequently, our forecasting objective is reformulated to map the conditional distribution $P(\mathbf{Y} \mid \mathbf{X}, \mathbf{R})$.

Our framework is fundamentally inspired by the architectural success of conditional diffusion models and knowledge distillation \citep{rombach2022high, zhang2023adding}. Intuitively, historical context $\mathbf{X}$ alone often admits a multimodal predictive distribution over divergent future trajectories. Projecting a distilled reasoning prior $\mathbf{R}$ directly into the model's continuous latent space disambiguates these modes, guiding the forecaster toward the target trajectory and substantially reducing predictive uncertainty (formalized under a variance-reduction framework in App.~\ref{app:theory}).

\subsection{Reasoning Generation and Distillation}
\label{subsec:data_gen}

\paragraph{Phase 1: Oracle Reasoning Generation.} First, we construct a high-quality reference reasoning dataset using the training splits of the GIFT-Eval \citep{aksu2025gifteval}, partitioning sampled temporal windows into a historical context $X$ and a pseudo-future ground truth $Y$. We adapt the oracle reasoning generation method introduced in TFRBench \citep{ahamed2026tfrbench} to generate target reasoning traces. Using a highly capable Teacher LLM (e.g., Gemini-3.1-Pro) as base model in multi-agent system, we retrieve time-bound external events $\mathcal{E}$ corresponding to the temporal span and generate a concrete, step-by-step oracle reasoning $R_{ref}$. This reasoning qualitatively details expected trend components, seasonality, and residual impacts based on cross-channel dependencies and $\mathcal{E}$.

\paragraph{Phase 2: Reasoning Distillation.} Next, we distill this reasoning capability into a lightweight student LLM. Crucially, while the teacher utilizes $Y$ and $\mathcal{E}$ to generate $R_{ref}$, the student is tasked with predicting equivalent analytical insights relying strictly on the historical context $X$. To bridge the modality gap, we serialize $X$ into a structured textual prompt $P_{in}$, augmented with global statistical metadata and a baseline reasoning draft. The student LLM processes $P_{in}$ to autoregressively generate a refined analytical strategy $\hat{R}$. We supervise this generation using standard causal language modeling, optimizing the Cross-Entropy loss $\mathcal{L}_{CE}$ exclusively over the generated reasoning tokens:
\begin{equation*}
\mathcal{L}_{CE} = -\sum_{j=1}^{M} \log P(R_{ref,j} | P_{in}, R_{ref,<j}; \Theta_{LLM})
\end{equation*}
where $M$ is the total token count of the reference reasoning and $\Theta_{LLM}$ represents the trainable parameters of the Student LLM. By masking the input prompt tokens during loss calculation, we structurally fine-tune the Student to articulate forward-looking insights while its prompt (pre-fill) states are projected into the time series space. Oracle access to $Y$ and $\mathcal{E}$ is used only on training windows. The student (and test-time inference) never receives future values or retrieved events. This prevents direct leakage of any future values.

\subsection{Cross-Modal Fusion and Training}

As formalized in Algorithm \ref{alg:method} (App.~\ref{app:algorithms}), we bypass discrete tokenization to inject the analytical strategy into the numerical forecaster via a continuous, end-to-end trainable fusion pipeline.

\paragraph{Step 1: Latent Projection.} Let $H \in \mathbb{R}^{L \times d_{llm}}$ represent the last hidden states of the Student LLM. We mean-pool these states over the sequence dimension to extract a global reasoning embedding $h_R$. A trainable linear projection matrix $W_{proj}$ maps $h_R$ to the exact dimensionality of the target TSFM, creating the cross-modal reasoning prior $e_R = h_R W_{proj}$.

\paragraph{Step 2: Generalized Latent Fusion.} The historical data $X$ is independently processed into native time series embeddings $E_{TS}$. We integrate the modalities using a generalized fusion operator: $E_{fused} = e_R \oplus E_{TS}$. To ensure architectural adaptability, $\oplus$ represents sequence prefixing for direct multi-step forecasters (e.g., Chronos-2.0) or initial state substitution to preserve sequence lengths in autoregressive forecasters (e.g., Timer-S1). The TSFM processes $E_{fused}$ to output predicted quantiles $\hat{Y}_q$, penalized by $\mathcal{L}_{Quantile}$.

\paragraph{Step 3: Joint Optimization.} The architecture is optimized simultaneously for language generation and forecasting using the combined objective:

\begin{equation}
    \mathcal{L}_{total} = \alpha \mathcal{L}_{CE} + \beta \mathcal{L}_{Quantile}
    \label{eq:total_loss}
\end{equation}

Gradients flow from the TSFM decoder, through the projection layer, and directly into the Student LLM. To preserve foundational linguistic knowledge while adapting to the forecasting task, we optimize the Student LLM exclusively via Low-Rank Adaptation (LoRA) on its attention layers.

\subsection{Evaluation Phase}
\label{subsec:evaluation}

During inference, \ours{} does not use future ground-truth $Y$ or the event-search agent; Algorithm~\ref{alg:inference} may only query a large LLM once strictly on historical $X$ to draft $R_{base}$. Given an unseen historical context $X$, we construct the prompt $P_{in}$. The student LLM processes this prompt to autoregressively generate the explicit semantic reasoning $\hat{R}$, providing practitioners with a human-interpretable strategic narrative for the forecast.

Concurrently, the Student LLM's continuous hidden states are mean-pooled, projected into the reasoning prior $e_R$, and fused with the native time series embeddings $E_{TS}$. Conditioned on this injected strategic guide, the foundation model computes the future trajectory, outputting predicted quantiles $\hat{Y}_q$ across a discrete set of target levels $\mathcal{Q}=\{0.1, 0.2, \dots, 0.9\}$. This dual-generation pipeline simultaneously yields both reasoning and continuous numerical forecasts. We provide the steps in more details in App.~\ref{app:algorithms}, Algorithm~\ref{alg:inference}.

\section{Experiments and Results}
% In this section, we present our experimental details and the results. 
\subsection{Experimental Setup}
\label{subsec:experimental_setup}

\paragraph{Datasets}
We evaluate our framework on two benchmarks. For continuous numerical forecasting precision, we utilize the GIFT-Eval benchmark \citep{aksu2025gifteval}. To assess further the generalizability of our framework we perform in-domain and out-of-domain forecasting as well as reasoning quality evaluation on TFRBench~\citep{ahamed2026tfrbench} with 10 diverse datasets from five different domains. These TFRBench datasets include both in-domain (i.e., tasks similar to GIFT-Eval training set) and out-of-domain evaluation sets. To avoid any seen data or leakage issue for in-domain evaluation on TFRBench, we excluded the samples, which are used during training of our framework.

\paragraph{Baselines}
We benchmark against traditional statistical methods, TSFMs, and LLMs (App.~\ref{app:baseline}), prioritizing TSFMs with reported GIFT-Eval and TFRBench results. PatchTST~\citep{nie2023a} is related but not a matched GIFT-Eval baseline, as it is generally trained per task. Task-specific TS-LLMs such as Time-LLM~\citep{jin2024timellm} are compared in Sec.~\ref{sec:ablation} on a multi-domain subset, since they are relevant but not pretrained-TSFM adapters under the GIFT-Eval protocol.

\paragraph{Evaluation Metrics}
To provide a comprehensive evaluation, we utilize both quantitative and qualitative metrics. For numerical forecasting accuracy, we report the Mean Absolute Scaled Error (MASE), Mean Absolute Error (MAE), and Continuous Ranked Probability Score (CRPS) to measure both point-forecast precision and distributional density estimation. For reasoning quality evaluation, motivated from \citet{ahamed2026tfrbench}, we employ an LLM-as-a-Judge protocol that scores the generated strategic narratives on a scale of 1 to 5 across four critical dimensions: Domain Relevance, Forecasting Correctness, Event Relevance, and Logic-to-Number Consistency.

\paragraph{Implementation Details}
Our pipeline utilizes Gemini-3.1-Pro to generate reference reasoning and optimizes a Gemma-3-4B-it student model (via LoRA) jointly with the TSFM backbone using a combined textual and numerical loss. Implementation specifics, including hyperparameters, hardware configurations, alternative model variants, and prompts are detailed in App.~\ref{app:implementation} and App.~\ref{app:prompts_llm}.

\subsection{Numerical Performance}

\subsubsection{Evaluation on GIFT-Eval}

\begin{figure}
    \centering
    \includegraphics[width=0.65\linewidth]{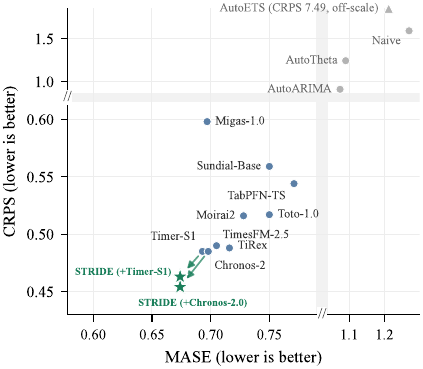}
    % \vspace{-0.25cm}
    \caption{GIFT-Eval benchmark results. Integrating our framework with models like Chronos-2.0 and Timer-S1 reduces forecasting error compared to standard baselines.  The cross-modal design grounds numerical predictions in a strategic analytical prior, consistently delivering highly robust forecasts.}
    \label{fig:gifteval_result}
\end{figure}
To evaluate the numerical forecasting capabilities and architectural generalizability of our approach, we benchmark it against state-of-the-art foundation models and traditional statistical methods on the GIFT-Eval leaderboard. To demonstrate that our framework is generalizable to the underlying architecture, we integrated it with two distinct Time Series Foundation Models (TSFMs): Chronos-2.0~\citep{ansari2025chronos} and Timer-S1~\citep{liu2026timer}. These specific models were selected because they represent recent state-of-the-art methodologies with accessible open-source code and fine-tunable checkpoints, unlike many proprietary alternatives that restrict required modifications.

As illustrated in Figure~\ref{fig:gifteval_result}, our approach establishes a new state-of-the-art, securing the top performance across both deterministic and probabilistic metrics. Specifically, our framework achieves a Mean Absolute Scaled Error (MASE) of 0.674 and a Continuous Ranked Probability Score (CRPS) of 0.454. Furthermore, to quantify the direct impact of our cross-modal projection, we compare the original numerical-only Timer-S1 baseline against our reasoning-augmented configuration. Integrating the language model's analytical strategy yields consistent improvements over the highly optimized base model, reducing the MASE from 0.693 to 0.674 and the CRPS from 0.485 to 0.463. 

These results indicate that the latent reasoning prior guides the TSFM toward more accurate forecasts and tighter predictive distributions. This validates that even when applied to mature foundation models, the explicit linguistic modeling of trend shifts and cross-channel dynamics provides a critical inductive bias that purely numerical encoders struggle to capture alone. Following standard practice from other works~\citep{liu2026timer}, we report the MASE and CRPS for GIFT-Eval.

\subsubsection{Evaluation on TFRBench}

\begin{table*}
\centering
\caption{Numerical performance on TFRBench. We report Mean Absolute Error (MAE) and Mean Absolute Scaled Error (MASE) (lower is better). To avoid cross-dataset scaling issues, we only present the average for MASE. \textbf{Bold}: best, and \underline{underline}: second best. The detailed analysis with standard deviations are shown in the App.~\ref{app:tfrbench_numerical}. App.~\ref{app:metric_uncertainty} reports metric uncertainty and a conservative $+1$-std check \ours{} \textit{vs.} Chronos-2.0.}
\resizebox{\linewidth}{!}{%
\begin{tabular}{@{}l|cc|cc|cc|cc|cc|c@{}}
\toprule
\multicolumn{12}{c}{\textbf{TFRBench (In-Domain)}} \\ \midrule
\multirow{2}{*}{\textbf{Models}} & \multicolumn{2}{c|}{\textbf{Solar Daily}} & \multicolumn{2}{c|}{\textbf{Electricity}} & \multicolumn{2}{c|}{\textbf{Car Parts}} & \multicolumn{2}{c|}{\textbf{Hierarchical Sales}} & \multicolumn{2}{c|}{\textbf{Bitbrains Fast Storage}} & \textbf{Average} \\ \cmidrule(l){2-12} 
 & \textbf{MAE} & \textbf{MASE} & \textbf{MAE} & \textbf{MASE} & \textbf{MAE} & \textbf{MASE} & \textbf{MAE} & \textbf{MASE} & \textbf{MAE} & \textbf{MASE} & \textbf{MASE} \\ \midrule
ARIMA & $1.444$ & $0.699$ & $5.564$ & $1.536$ & $0.569$ & $0.873$ & $2.239$ & $0.789$ & $3.57 \times 10^{4}$ & $0.799$ & $0.939$ \\
TimesFM-2.5 & $1.475$ & $0.735$ & $4.758$ & $1.380$ & $0.328$ & $0.448$ & $2.142$ & $0.753$ & $\underline{2.62 \times 10^{4}}$ & $3.401$ & $1.343$ \\
Chronos-2.0 & $\underline{1.409}$ & $\underline{0.685}$ & $4.162$ & $1.271$ & $\underline{0.294}$ & $\underline{0.413}$ & $\underline{2.127}$ & $\underline{0.750}$ & $2.69 \times 10^{4}$ & $\underline{0.708}$ & $\underline{0.765}$ \\
\midrule
Gemini-2.5-Flash & $2.166$ & $1.037$ & $6.271$ & $1.785$ & $0.460$ & $0.662$ & $2.675$ & $0.904$ & $4.62 \times 10^{4}$ & $1.110$ & $1.100$ \\
Gemini-2.5-Pro & $2.598$ & $1.217$ & $7.604$ & $2.244$ & $0.432$ & $0.630$ & $2.775$ & $0.938$ & $4.99 \times 10^{4}$ & $1.266$ & $1.259$ \\
Claude-Sonnet-4 & $2.481$ & $1.169$ & $8.774$ & $2.647$ & $0.475$ & $0.703$ & $3.417$ & $1.125$ & $1.11 \times 10^{5}$ & $3.026$ & $1.734$ \\
Claude-Sonnet-4.5 & $1.780$ & $0.869$ & $6.316$ & $1.714$ & $0.413$ & $0.595$ & $2.710$ & $0.898$ & $8.34 \times 10^{4}$ & $3.244$ & $1.464$ \\
Gemini-3.1-Pro & $1.603$ & $0.779$ & $\underline{3.944}$ & $\underline{1.184}$ & $0.476$ & $0.671$ & $2.341$ & $0.801$ & $4.16 \times 10^{4}$ & $0.833$ & $0.854$ \\
\ours{} (+Chronos-2.0)& $\mathbf{0.884}$ & $\mathbf{0.448}$ & $\mathbf{2.922}$ & $\mathbf{0.828}$ & $\mathbf{0.280}$ & $\mathbf{0.404}$ & $\mathbf{2.081}$ & $\mathbf{0.719}$ & $\mathbf{2.47 \times 10^{4}}$ & $\mathbf{0.676}$ & $\mathbf{0.615}$ \\
\midrule
\multicolumn{12}{c}{\textbf{TFRBench (Out-of-Domain)}} \\ \midrule
\multirow{2}{*}{\textbf{Models}} & \multicolumn{2}{c|}{\textbf{Web Traffic}} & \multicolumn{2}{c|}{\textbf{Traffic}} & \multicolumn{2}{c|}{\textbf{Nyc Taxi}} & \multicolumn{2}{c|}{\textbf{Amazon}} & \multicolumn{2}{c|}{\textbf{Apple}} & \textbf{Average} \\ \cmidrule(l){2-12} 
 & \textbf{MAE} & \textbf{MASE} & \textbf{MAE} & \textbf{MASE} & \textbf{MAE} & \textbf{MASE} & \textbf{MAE} & \textbf{MASE} & \textbf{MAE} & \textbf{MASE} & \textbf{MASE} \\ \midrule
ARIMA & $15.400$ & $0.788$ & $0.039$ & $2.628$ & $8.65 \times 10^{3}$ & $3.429$ & $9.11 \times 10^{6}$ & $0.995$ & $2.28 \times 10^{7}$ & $1.100$ & $1.788$ \\
TimesFM-2.5 & $\underline{12.000}$ & $\underline{0.676}$ & $0.014$ & $0.847$ & $2.70 \times 10^{3}$ & $1.051$ & $\mathbf{6.56 \times 10^{6}}$ & $\mathbf{0.747}$ & $\mathbf{1.86 \times 10^{7}}$ & $\mathbf{0.879}$ & $0.840$ \\
Chronos-2.0 & $\mathbf{11.900}$ & $0.679$ & $\underline{0.011}$ & $\underline{0.664}$ & $1.96 \times 10^{3}$ & $\underline{0.790}$ & $7.24 \times 10^{6}$ & $0.821$ & $\underline{1.95 \times 10^{7}}$ & $\underline{0.936}$ & $\underline{0.778}$ \\
\midrule
Gemini-2.5-Flash & $23.300$ & $1.120$ & $0.027$ & $1.647$ & $5.75 \times 10^{3}$ & $2.289$ & $1.37 \times 10^{7}$ & $1.333$ & $3.37 \times 10^{7}$ & $1.482$ & $1.574$ \\
Gemini-2.5-Pro & $28.200$ & $1.538$ & $0.030$ & $1.885$ & $7.10 \times 10^{3}$ & $2.815$ & $1.49 \times 10^{7}$ & $1.464$ & $3.57 \times 10^{7}$ & $1.589$ & $1.858$ \\
Claude-Sonnet-4 & $19.900$ & $1.211$ & $0.024$ & $1.450$ & $5.31 \times 10^{3}$ & $2.085$ & $1.34 \times 10^{7}$ & $1.511$ & $3.54 \times 10^{7}$ & $1.651$ & $1.582$ \\
Claude-Sonnet-4.5 & $18.000$ & $0.865$ & $0.024$ & $1.458$ & $4.63 \times 10^{3}$ & $1.837$ & $7.97 \times 10^{6}$ & $0.870$ & $2.16 \times 10^{7}$ & $0.989$ & $1.204$ \\
Gemini-3.1-Pro & $13.600$ & $1.472$ & $0.015$ & $1.102$ & $\underline{1.73 \times 10^{3}}$ & $0.889$ & $1.03 \times 10^{7}$ & $1.776$ & $2.70 \times 10^{7}$ & $1.802$ & $1.408$ \\
\ours{} (+Chronos-2.0)& $12.076$ & $\mathbf{0.664}$ & $\mathbf{0.011}$ & $\mathbf{0.634}$ & $\mathbf{1.39 \times 10^{3}}$ & $\mathbf{0.555}$ & $\underline{6.56 \times 10^{6}}$ & $\underline{0.791}$ & $1.99 \times 10^{7}$ & $0.973$ & $\mathbf{0.724}$ \\
\bottomrule
\end{tabular}
}
\label{tab:numerical_tfrbench_combined}
\end{table*}
To rigorously assess the generalizability of our framework, we conduct an in-domain and out-of-domain evaluation using ten diverse datasets from TFRBench whose evaluation windows were excluded from our training. As detailed in Table~\ref{tab:numerical_tfrbench_combined}, our approach achieves the best overall performance, recording the lowest average Mean Absolute Scaled Error (MASE) of 0.615 for in-domain and 0.724 for out-of-domain datasets. Notably, the framework tightens error margins on complex series such as NYC Taxi and Traffic, outperforming dedicated numerical encoders like TimesFM-2.5 and Chronos-2.0. By injecting the analytical strategy as a continuous latent prior, our method overcomes the forecasting limitations inherent in purely text-based LLMs like Gemini-3.1-Pro, demonstrating that explicit reasoning serves as a powerful, transferable inductive bias for forecasting tasks.

\subsection{Reasoning Performance on TFRBench}
\label{subsec:tfrbench_reasoning}
\begin{figure}
    \centering
    \includegraphics[width=\linewidth]{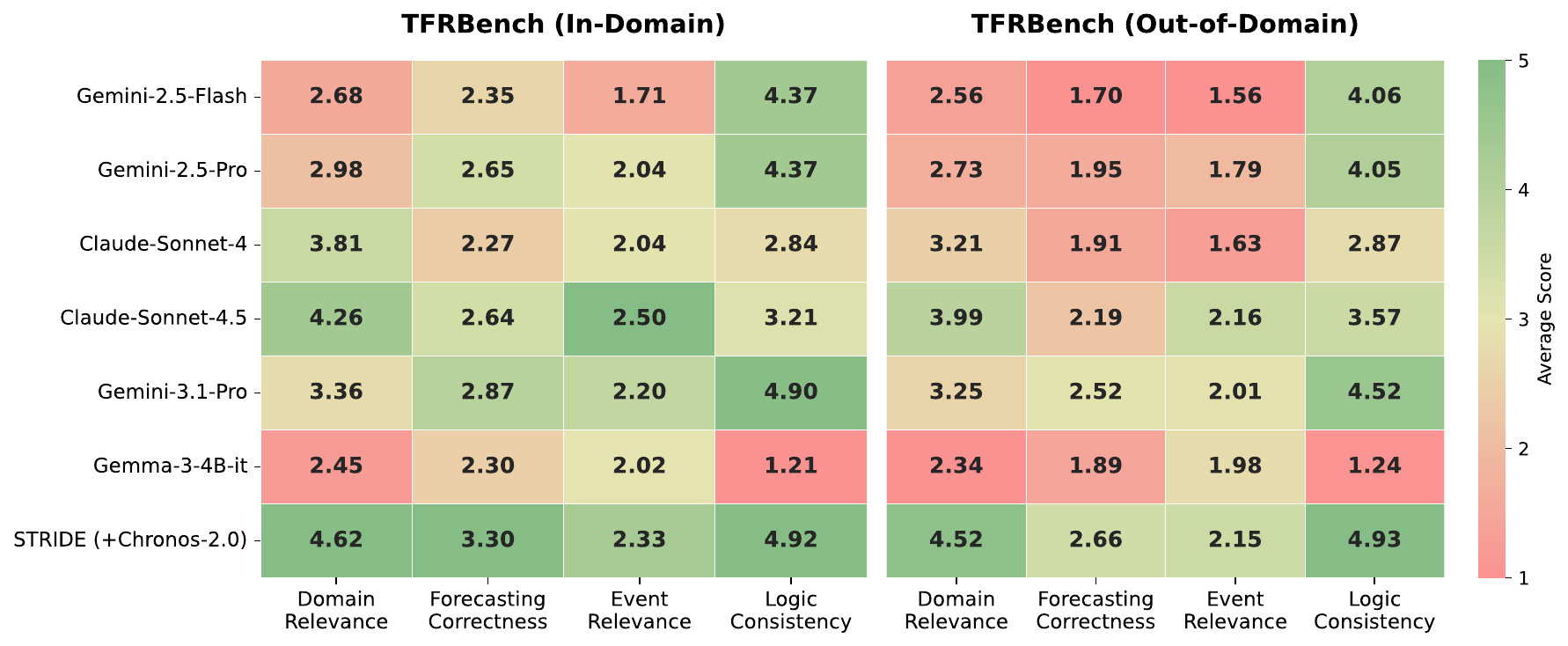}
    % \vspace{-0.4cm}
    \caption{Reasoning performance on TFRBench datasets. Scores are measured on a 1–5 scale, where higher values indicate superior reasoning performance. Detailed results are shown in App.~\ref{app:tfrbench_all_reason}.}
    \label{fig:tfrbench_reason_result}
\end{figure}

While quantitative metrics capture forecasting precision, the interpretability and logical coherence of the generated analytical strategies are equally paramount. To evaluate the generalization of our framework's textual reasoning capabilities, we conducted both in-domain and out-of-domain assessment using the TFRBench datasets, whose evaluation windows were held out from training.

We employ an LLM-as-a-Judge protocol to score the generated narratives across four critical dimensions on a scale of 1 to 5: Domain Relevance, Forecasting Correctness, Event Relevance, and Logic-to-Number Consistency. App.~\ref{app:human_eval_tfrbench} summarizes human-calibrated LLM-as-a-Judge protocol. Figure~\ref{fig:tfrbench_reason_result} presents the aggregated average scores across in-domain and out-of-domain datasets. The generated outputs are formatted for evaluation using Gemini-3.1-Pro to ensure a structured style.

Our framework consistently achieves the highest average performance across majority of the evaluation criteria. The strong performance in Forecasting Correctness, Domain Relevance and  Logic Consistency indicates that our framework successfully grounds its reasoning in actual domain dynamics, effectively avoiding the hallucinations and logical disconnects often observed when utilizing standard foundational LLMs for time series tasks. We also provide several case studies in App.~\ref{app:forecasting_case} and App.~\ref{app:reasoning_case} for forecasting and reasoning respectively. The detailed, dataset-by-dataset breakdown of these reasoning evaluation scores can be found in App.~\ref{app:tfrbench_all_reason} (Table~\ref{tab:judge-performance-indomain} and \ref{tab:judge-performance-ood}). Furthermore, \textsection~\ref{app:reasoning_mase_correlation} further confirms the correlation that higher reasoning quality consistently yields reduced MASE.

\section{Discussion}
\label{sec:discussion}

Here we discuss where the gains of \ours{} come from and what they imply for reasoning-aware forecasting. We first present ablations that isolate structured reasoning from capacity and architecture choices (\textsection ~\ref{sec:ablation}), then summarize the resulting findings (\textsection ~\ref{sec:findings}).

\subsection{Ablations}
\label{sec:ablation}

To determine whether aligned, structured reasoning drives predictive accuracy, we substitute alternative student and teacher LLM backbones to assess cross-architecture generalizability. We then evaluate numerical fine-tuning without reasoning and shuffled series--reasoning pairings (Table~\ref{tab:chronos_ft_control}), text-to-number TS-LLM baselines (Table~\ref{tab:tsllm_baselines}), and generic summary embeddings (App.~\ref{app:additional_ablations}) to rule out potential confounds and validate our cross-modal design from multiple angles.

\begin{figure}[t]
    \centering
    \begin{subfigure}[t]{\linewidth}
        \centering
        \includegraphics[width=0.75\linewidth]{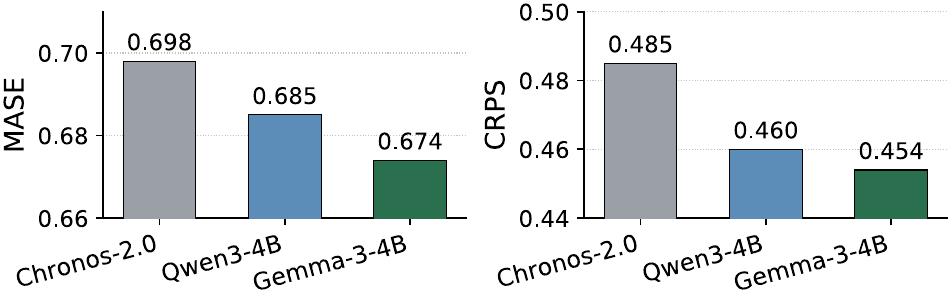}
        \centering
        \subcaption{\centering Student (Small-LLM) variants.}
        \label{fig:ablation_small_llm}
    \end{subfigure}
    \begin{subfigure}[t]{\linewidth}
        \centering
        \includegraphics[width=0.75\linewidth]{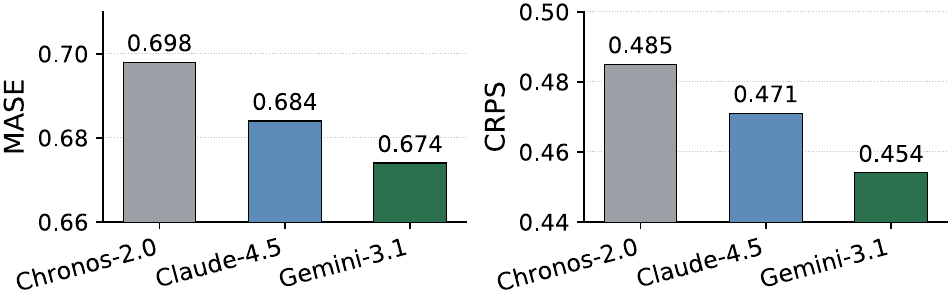}
        \centering
        \subcaption{\centering Baseline-reasoning provider (Large-LLM) variants.}
        \label{fig:ablation_large_llm}
    \end{subfigure}
    \caption{Ablation of the LLMs inside \ours{} (+Chronos-2.0) on GIFT-Eval (aggregate MASE/CRPS; lower is better): (a) the distilled student, (b) the large LLM that drafts the baseline reasoning. Every variant beats the unaugmented Chronos-2.0 (0.698/0.485); the default pair (Gemma-3-4B-it, Gemini-3.1-Pro) is best (0.674/0.454), so the gains are not tied to one LLM.}
    \label{fig:ablation_llm_variants}
\end{figure}
\subsubsection{LLM Variants}
\label{app:llm_variants}
\paragraph{Small LLM Variants.}
\label{subsec:small_llm}
To evaluate the robustness of the knowledge distillation and cross-modal projection, we conducted an ablation study substituting the student model. During optimization, the Small-LLM generates the reasoning strategy while providing its mean-pooled hidden states as a continuous prior to the numerical encoder. Figure~\ref{fig:ablation_small_llm} compares GIFT-Eval performance using Gemma-3-4B-it versus Qwen3-4B-Instruct against the numerical Chronos-2.0 baseline. The framework remains highly effective across different student architectures. While Gemma-3-4B-it achieves the lowest errors (MASE of 0.674, CRPS of 0.454), Qwen3-4B-Instruct also yields highly competitive results surpassing the unaugmented baseline. 

\paragraph{Large LLM Variants.}
\label{subsec:large_llm}

Figure~\ref{fig:ablation_large_llm} confirms \ours{}'s improvements are not tied to a specific baseline LLM. Training the student (Gemma-3-4B-it) with reasoning from either Gemini-3.1-Pro or Claude-Sonnet-4.5 consistently outperforms the numerical Chronos-2.0 baseline (e.g., Claude reduces MASE from 0.698 to 0.684). This demonstrates explicit reasoning is a robust inductive bias, allowing diverse LLMs to bridge the modality gap.

\subsubsection{No-Reasoning Fine-Tune and Pairing}
\label{subsec:additional_controls}
Fine-tuning Chronos-2.0 on the same windows without a student LLM or reasoning traces (Table~\ref{tab:chronos_ft_control}; Details in App.~\ref{app:chronos_ft_control}) improves over regular Chronos-2.0 (e.g., Solar MASE $1.012 \to 0.996$; BizITObs L2C $0.647 \to 0.523$), yet remains worse than \ours{} ($0.315$ and $0.429$). A shuffled ablation that deranges reasoning across samples (Table~\ref{tab:train_pairing_shuffle}; Details in App.~\ref{app:train_pairing_shuffle}) is higher-error on the datasets (e.g., Solar $0.985$ vs.\ $0.315$; US Births $0.834$ vs.\ $0.627$). Thus, neither fine-tuning the TSFM on the series alone nor feeding it reasoning from a different sample accounts for the gains of \ours{}.

\begin{table*}[t]
\centering
\footnotesize
\caption{No-reasoning Chronos-2.0 fine-tuning and paired vs.\ shuffled reasoning on GIFT-Eval datasets with varying frequencies. Paired is \ours{}. We report MAE and MASE (lower is better). \textbf{Bold}: best; \underline{underline}: second best.}
\label{tab:chronos_ft_control}
\label{tab:train_pairing_shuffle}
\resizebox{\linewidth}{!}{%
\begin{tabular}{lll *{2}{c} *{2}{c} *{2}{c} *{2}{c}}
\toprule
\multirow{2}{*}{\textbf{Dataset}} & \multirow{2}{*}{\textbf{Domain}} & \multirow{2}{*}{\textbf{Freq.}} &
\multicolumn{2}{c}{\textbf{Chronos-2.0}} &
\multicolumn{2}{c}{\textbf{Chronos-2.0 FT}} &
\multicolumn{2}{c}{\textbf{Shuffled}} &
\multicolumn{2}{c}{\textbf{\ours{} (Paired)}} \\
\cmidrule(lr){4-5}\cmidrule(lr){6-7}\cmidrule(lr){8-9}\cmidrule(lr){10-11}
& & &
\textbf{MAE} & \textbf{MASE} &
\textbf{MAE} & \textbf{MASE} &
\textbf{MAE} & \textbf{MASE} &
\textbf{MAE} & \textbf{MASE} \\
\midrule
Solar & Energy & W &
$9.317\!\times\!10^{2}$ & $1.012$ &
$9.146\!\times\!10^{2}$ & $0.996$ &
$\underline{9.021\!\times\!10^{2}}$ & $\underline{0.985}$ &
$\mathbf{2.955\!\times\!10^{2}}$ & $\mathbf{0.315}$ \\
BizITObs L2C & CloudOps & 5T &
$6.571\!\times\!10^{0}$ & $0.647$ &
$5.250\!\times\!10^{0}$ & $0.523$ &
$\underline{4.358\!\times\!10^{0}}$ & $\underline{0.431}$ &
$\mathbf{4.321\!\times\!10^{0}}$ & $\mathbf{0.429}$ \\
KDD Cup 2018 & Nature & H &
$2.479\!\times\!10^{1}$ & $1.047$ &
$2.480\!\times\!10^{1}$ & $1.045$ &
$\underline{1.904\!\times\!10^{1}}$ & $\underline{0.832}$ &
$\mathbf{1.662\!\times\!10^{1}}$ & $\mathbf{0.721}$ \\
US Births & Healthcare & M &
$5.613\!\times\!10^{3}$ & $0.633$ &
$\underline{5.598\!\times\!10^{3}}$ & $\underline{0.631}$ &
$1.056\!\times\!10^{4}$ & $0.834$ &
$\mathbf{5.532\!\times\!10^{3}}$ & $\mathbf{0.627}$ \\
Hierarchical Sales & Sales & D &
$2.311\!\times\!10^{0}$ & $0.744$ &
$\underline{2.301\!\times\!10^{0}}$ & $\underline{0.743}$ &
$2.564\!\times\!10^{0}$ & $0.816$ &
$\mathbf{2.273\!\times\!10^{0}}$ & $\mathbf{0.738}$ \\
\bottomrule
\end{tabular}%
}
\end{table*}

\begin{table*}
\centering
% \footnotesize
\caption{TS-LLM baselines vs.\ \ours{}. MAE and MASE (lower is better). \textbf{Bold}: best; \underline{underline}: second best.}
\label{tab:tsllm_baselines}
\resizebox{\linewidth}{!}{%
\begin{tabular}{lll *{2}{c} *{2}{c} *{2}{c} *{2}{c}}
\toprule
\multirow{2}{*}{\textbf{Dataset}} & \multirow{2}{*}{\textbf{Domain}} & \multirow{2}{*}{\textbf{Freq.}} &
\multicolumn{2}{c}{\textbf{Time-LLM}} &
\multicolumn{2}{c}{\textbf{PromptCast}} &
\multicolumn{2}{c}{\textbf{Chow CoT}} &
\multicolumn{2}{c}{\textbf{\ours{}}} \\
\cmidrule(lr){4-5}\cmidrule(lr){6-7}\cmidrule(lr){8-9}\cmidrule(lr){10-11}
& & &
\textbf{MAE} & \textbf{MASE} &
\textbf{MAE} & \textbf{MASE} &
\textbf{MAE} & \textbf{MASE} &
\textbf{MAE} & \textbf{MASE} \\
\midrule
Solar & Energy & W &
$\underline{1.657\!\times\!10^{3}}$ & $\underline{1.823}$ &
$3.062\!\times\!10^{3}$ & $3.370$ &
$3.137\!\times\!10^{3}$ & $3.452$ &
$\mathbf{2.955\!\times\!10^{2}}$ & $\mathbf{0.315}$ \\
BizITObs L2C & CloudOps & 5T &
$1.471\!\times\!10^{1}$ & $1.418$ &
$\underline{1.367\!\times\!10^{1}}$ & $\underline{1.316}$ &
$1.382\!\times\!10^{1}$ & $1.330$ &
$\mathbf{4.321\!\times\!10^{0}}$ & $\mathbf{0.429}$ \\
KDD Cup 2018 & Nature & H &
$\underline{2.588\!\times\!10^{1}}$ & $\underline{1.251}$ &
$6.585\!\times\!10^{1}$ & $2.742$ &
$1.177\!\times\!10^{2}$ & $4.902$ &
$\mathbf{1.662\!\times\!10^{1}}$ & $\mathbf{0.721}$ \\
US Births & Healthcare & M &
$8.417\!\times\!10^{4}$ & $9.517$ &
$2.531\!\times\!10^{4}$ & $2.862$ &
$\underline{1.123\!\times\!10^{4}}$ & $\underline{1.270}$ &
$\mathbf{5.532\!\times\!10^{3}}$ & $\mathbf{0.627}$ \\
Hierarchical Sales & Sales & D &
$\underline{2.579\!\times\!10^{0}}$ & $\underline{0.859}$ &
$3.412\!\times\!10^{0}$ & $1.135$ &
$5.313\!\times\!10^{0}$ & $1.768$ &
$\mathbf{2.273\!\times\!10^{0}}$ & $\mathbf{0.738}$ \\
\bottomrule
\end{tabular}%
}
\end{table*}

\subsubsection{Generic Summary Embeddings and TS-LLMs}

Replacing student hidden states with a matched-dimension non-strategy summary reasoning through the same $W_{\mathrm{proj}}$ (App.~\ref{app:generic_summary_control}) does not match \ours{} (Solar MASE $1.270$ vs.\ $0.315$; KDD Cup $1.093$ vs.\ $0.721$) on majority of the datasets. Besides, Time-LLM, PromptCast, and Chow-style CoT remain higher-error in MAE and MASE (Table~\ref{tab:tsllm_baselines}; Details in App.~\ref{app:tsllm_baselines}). An arbitrary extra channel, or decoding numbers from text tokens, is not a substitute for \ours{}: the gains require correctly paired, strategy-bearing reasoning rather than capacity alone.

\subsubsection{Compatibility with Ensembling}
\label{app:ensemble}
Several of the strongest GIFT-Eval entries are ensembles of TSFMs, e.g., Synapse~\citep{das2026synapse}. To test the generalizability of \ours{} in this setting, we evaluate its compatibility with both a generic mean ensemble and an advanced ensembling method. Specifically, we consider (i) Synapse without and with \ours{}, and (ii) a mean ensemble that averages the predicted quantiles of Chronos-2.0 and Timer-S1 versus those of \ours{} (+Chronos-2.0) and \ours{} (+Timer-S1), scored with the standard GIFT-Eval protocol.

\begin{figure}[H]
\centering
\includegraphics[width=0.5\linewidth]{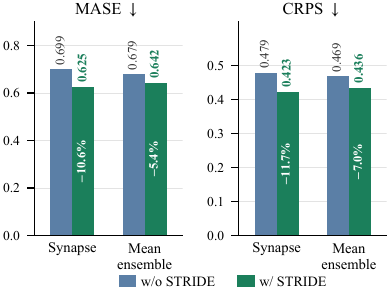}
\caption{Compatibility of \ours{} with ensembling on GIFT-Eval (aggregate MASE and CRPS; lower is better). \emph{Synapse}: the public-leaderboard Synapse ensemble without and with \ours{}. \emph{Mean ensemble}: mean of the quantile forecasts of Chronos-2.0 and Timer-S1 (w/o \ours{}) versus the mean of \ours{} (+Chronos-2.0) and \ours{} (+Timer-S1) (w/ \ours{}). Percentages inside the bars are the relative reduction from adding \ours{}.}
\label{fig:ensemble_gifteval}
\end{figure}

As shown in Figure~\ref{fig:ensemble_gifteval}, \ours{} improves both ensembles: Synapse goes from $0.699/0.479$ to $0.625/0.423$ (MASE/CRPS), a relative reduction of $10.6\%$ and $11.7\%$, and the mean ensemble from $0.679/0.469$ to $0.642/0.436$, a reduction of $5.4\%$ and $7.0\%$. As ensembling also adds stronger backbones to the pool, we view these results as evidence that \ours{} is compatible with, and complementary to, multiple ensembling methods.

\subsubsection{Reasoning Quality vs.\ Forecasting Error}
\label{app:reasoning_mase_correlation}
\begin{figure}
\centering
\includegraphics[width=0.8\linewidth]{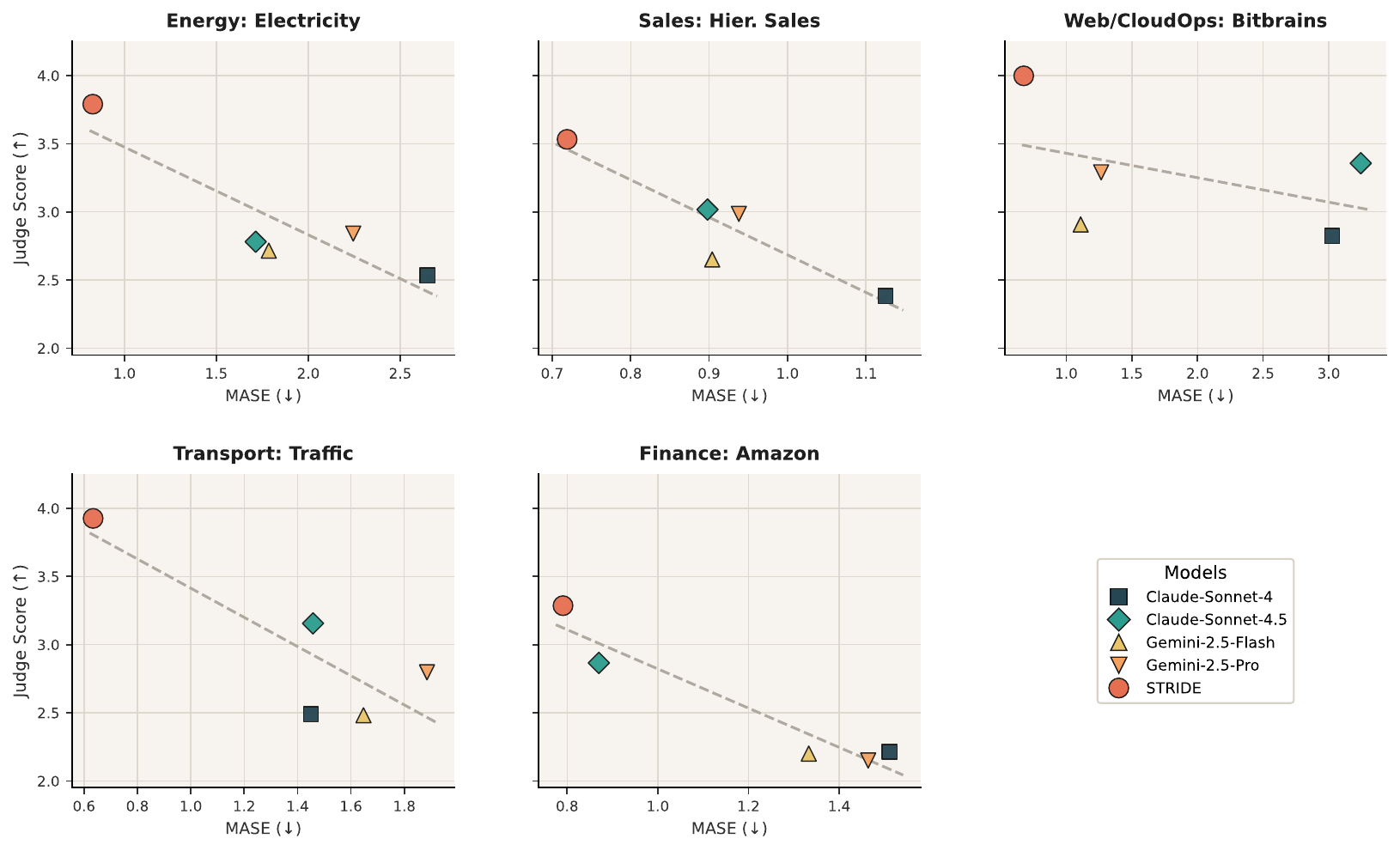}
\caption{Judge Score vs.\ MASE on one TFRBench dataset per domain. Each point uses reported table means (Judge Score $=$ average of Dom./Fcst./Evt./Logic). \ours{} combines higher reasoning quality with lower forecast error relative to LLM baselines.}
\label{fig:reasoning_mase_corr}
\end{figure}

We further examine how qualitative reasoning scores relate to numerical forecast accuracy on TFRBench. For one representative dataset in each domain, we plot Judge Score against MASE for the LLM forecasters (Claude-Sonnet-4, Claude-Sonnet-4.5, Gemini-2.5-Flash, and Gemini-2.5-Pro) and for \ours{}. Judge Score is the mean of Domain Relevance, Forecasting Correctness, Event Relevance, and Logic-to-Number Consistency from TFRBench~\citep{ahamed2026tfrbench}. Figure~\ref{fig:reasoning_mase_corr} shows that \ours{} consistently attains both higher reasoning quality and lower MASE than the compared LLM baselines across all five domains.

\subsection{Findings}
\label{sec:findings}
The empirical results across the GIFT-Eval and TFRBench benchmarks, combined with our ablation studies, provide insights into why integrating language model reasoning improves numerical forecasting. Rather than only explanations, our framework demonstrates that explicit reasoning acts as a structural anchor. Several key findings highlight the strengths of this approach:
\vspace{-.4cm}
\paragraph{Reasoning Acts as a Directive for Forecast.} Our evaluation on the GIFT-Eval benchmark (Figure~\ref{fig:gifteval_result}) establishes that semantic reasoning serves as a prior that tightens distributional accuracy. The improvements in CRPS (0.454) and MASE (0.674) occur because the embedding information acts as a directive for prediction. By projecting this analytical strategy into the latent space, the model is structurally guided toward the future distribution mode, actively reducing predictive error before the numerical decoding step. This is consistent with our theoretical foundation. Since CRPS is a proper score of the forecast distribution, we do not observe $\mathrm{Var}(Y)$ directly. Attaching \ours{} reduces CRPS from $0.485$ to $0.454$ with Chronos-2.0 and from $0.485$ to $0.463$ with Timer-S1, matching a tighter predictive distribution than $P(Y \mid X)$. The no-reasoning, shuffle, and generic-summary controls further support the mode-selection assumption.

\paragraph{Resolving the Modality Gap.} Forcing continuous temporal data through discrete tokenizers degrades mathematical precision and introduces autoregressive bottlenecks (e.g., LLMTime; \citealp{gruver2024llmtime}). \ours{} resolves this by decoupling the modalities. Rather than generating numerical forecasts token-by-token, our framework extracts the mean-pooled hidden states from the Small-LLM's pre-fill phase (contextualizing the prompt and $R_{base}$). This continuous prior is projected into the TSFM to compute the forecast in a single forward pass. Concurrently, the Small-LLM decodes the refined reasoning ($\hat{R}$). This parallel execution avoids forcing LLMs into sequential numerical generation, structurally aligning the semantic narrative with the numerical output.

\paragraph{Generalizability.} Our ablation studies validate that the performance gains of \ours{} are not tied to a specific model family. \ours{} integrates with multiple foundation models, yielding consistent improvements over optimized baselines like Timer-S1 (reducing MASE from 0.693 to 0.674). This validates that explicit linguistic modeling provides a structural signal that numerical encoders simply cannot learn from raw arrays alone. Furthermore, we demonstrate the generalizability of \ours{} across multiple ensemble methods, showing further performance improvements for both generic mean and advanced ensembles (\textsection~\ref{app:ensemble}).

\paragraph{Eradicating Hallucinations.} By substituting various models (Figure~\ref{fig:ablation_llm_variants}), we verified that small LLM models can generate helpful strategic directives when trained using our framework, overcoming their original poor performance (Gemma-3-4B-it in Figure~\ref{fig:tfrbench_reason_result}). Furthermore, the reasoning evaluations for Logic-to-Number Consistency shows that our cross-modal projection enforces strict alignment between the generated narrative and the numerical output, validating that \ours{} provides robust inductive bias for forecasting.
\section{Conclusions and Future Work}
\label{sec:conclusion}

In this work, we introduced \ours{}, a framework bridging qualitative reasoning and continuous numerical forecasting. By bypassing discrete tokenization and dynamically projecting a language model's hidden states into a TSFM's embedding space, we natively equip numerical extrapolators with strategic planning. \ours{} achieves state-of-the-art precision on GIFT-Eval and TFRBench while generating accurate, logic-grounded narratives. Ablations confirm it serves as a robust, plug-and-play enhancement for foundational architectures like Chronos-2.0 and Timer-S1. Future research could extend this cross-modal distillation to other continuous modalities, such as spatial-temporal modeling or multi-agent reinforcement learning. By delivering interpretable narratives alongside precise predictions, \ours{} supports responsible deployment in high-stakes domains where transparency is critical. A discussion of the framework's limitations is provided in \textsection\ref{sec:limitations}.

\section{Limitations}
\label{sec:limitations}
While our approach effectively bridges the modality gap between semantic reasoning and numerical forecasting, its training pipeline is inherently bounded by the quality and factual accuracy of the Teacher (TFRBench~\citep{ahamed2026tfrbench}) used to generate the reference reasoning. Furthermore, jointly optimizing a Small-LLM alongside a target Time Series Foundation Model (TSFM) introduces additional computational overhead and memory requirements compared to deploying a standalone numerical encoder. 

\bibliographystyle{abbrvnat}
\bibliography{references}

%%%%%%%%%%%%%%%%%%%%%%%%%%%%%%%%%%%%%%%%%%%%%%%%%%%%%%%%%%%%

\appendix

\newpage

\clearpage
 % Switch to full-width (spans both columns)

% Define the visual style for the automated Appendix ToC
\titlecontents{section}[1.5em]
  {\vspace{0.5em}\bfseries\color{blue}}
  {\contentslabel{1.5em}}
  {\hspace*{-1.5em}}
  {\titlerule*[0.5pc]{.}\contentspage}

\titlecontents{subsection}[3.8em]
  {\normalfont\color{blue}}
  {\contentslabel{2.3em}}
  {\hspace*{-2.3em}}
  {\titlerule*[0.5pc]{.}\contentspage}

% Print the header
\vspace{2em}
\noindent{\Large\textbf{\textsf{Table of Contents for Appendix}}}\par
\vspace{1em}

% Start tracking sections and print the dynamic list
\startcontents[appendix]
\printcontents[appendix]{}{1}{\setcounter{tocdepth}{2}}
\vspace{2em}

\clearpage
 % Switch back to two columns for the following content

%%%%%%%%%%%%%%%%%%%%%%%%%%%%%%%%%
\section{Theoretical Intuition and Variance Proofs}
\label{app:theory}

This section details the theoretical framework, operational assumptions, cross-modal projection mechanism, and variance reduction properties underpinning the \ours{} architecture.

\subsection{Problem Setup and Assumptions}

\paragraph{Multi-Modality of Unconditional Forecasts.} Given historical context $\mathbf{X} \in \mathbb{R}^{T \times V}$, the true unconditional future distribution often exhibits multimodality due to unobserved exogenous factors or multiple regime shifts. We formalize this by modeling the future trajectory distribution as a mixture of $K$ distinct plausible trajectories (modes):
\begin{equation}
P(\mathbf{Y} \mid \mathbf{X}) = \sum_{i=1}^K \pi_i P_i(\mathbf{Y} \mid \mathbf{X}),
\end{equation}

where $P_i(\mathbf{Y} \mid \mathbf{X}) = \mathcal{N}(\boldsymbol{\mu}_i, \sigma_i^2 \mathbf{I})$ denotes the conditional distribution under mode $i$, and $\pi_i \ge 0$ ($\sum_{i=1}^K \pi_i = 1$) represents the historical prior probability of trajectory $i$.

\paragraph{Reasoning as a Latent Mode Selector.} In our distillation formulation, the oracle Teacher LLM generates reference reasoning $R_{\text{ref}}$ with access to privileged temporal events, successfully isolating the true future mode $k \in \{1, \dots, K\}$. The Student LLM processes $\mathbf{X}$ to output last hidden states $\mathbf{H} \in \mathbb{R}^{L \times d_{\text{llm}}}$. We operate under the regime where minimizing the textual cross-entropy loss $\mathcal{L}_{\text{CE}}$ aligns the mean-pooled representation $\mathbf{h}_R = \frac{1}{L}\sum_{l=1}^L \mathbf{H}_l$ with the oracle directive, such that:

\begin{equation}
P(\mathbf{Y} \mid \mathbf{X}, \mathbf{h}_R) \approx P_k(\mathbf{Y} \mid \mathbf{X}).
\end{equation}

\subsection{Mechanism of Cross-Modal Alignment}

Let $\mathbf{h}_R \in \mathbb{R}^{d_{\text{llm}}}$ be the sequence-averaged global reasoning state from the Student LLM. To map the semantic representation into the continuous temporal latent space, we introduce a learned linear projection matrix $\mathbf{W}_{\text{proj}} \in \mathbb{R}^{d_{\text{llm}} \times d_{\text{ts}}}$:
\begin{equation}
\mathbf{e}_R = \mathbf{h}_R \mathbf{W}_{\text{proj}}.
\end{equation}

The continuous reasoning prior $\mathbf{e}_R$ is fused with the native time series embeddings $\mathbf{E}_{\text{TS}} = \text{TSFM\_Embed}(\mathbf{X})$ via a generalized fusion operator $\oplus$ (e.g., sequence prefixing or initial-state injection):
\begin{equation}
\mathbf{E}_{\text{fused}} = \mathbf{e}_R \oplus \mathbf{E}_{\text{TS}}.
\end{equation}

The target TSFM decodes $\mathbf{E}_{\text{fused}}$ into the final forecast distribution $P(\mathbf{Y} \mid \mathbf{E}_{\text{fused}})$.

\subsection{Variance Reduction via Reasoning Injection}

\begin{proposition}[Variance Reduction]
\label{prop:variance_reduction}

Let $\mathbf{X}$ be the historical series and $\mathbf{E}_{\text{fused}}$ be the latent representation conditioning the TSFM on the aligned reasoning prior $\mathbf{e}_R$. Under the mixture model above, if modes share a common internal dispersion $\sigma^2$ and the reasoning prior correctly isolates the true future mode $k$, the predictive variance satisfies:
\begin{equation}
\mathrm{Var}(\mathbf{Y} \mid \mathbf{E}_{\text{fused}}) \le \mathrm{Var}(\mathbf{Y} \mid \mathbf{X}),
\end{equation}

with strict inequality holding whenever $K > 1$ and the mixture means $\boldsymbol{\mu}_i$ are not all identical.
\end{proposition}

\paragraph{Proof.}
Let $\bar{\boldsymbol{\mu}} = \sum_{i=1}^K \pi_i \boldsymbol{\mu}_i$ denote the unconditional expected future trajectory under past observations $\mathbf{X}$. 

\textit{Step 1 (Unconditional Variance):} By the law of total variance applied over the discrete mixture components:
\begin{align}
\mathrm{Var}(\mathbf{Y} \mid \mathbf{X}) &= \mathbb{E}_{i}[\mathrm{Var}(\mathbf{Y} \mid \mathbf{X}, i)] \nonumber \\
&\quad + \mathrm{Var}_{i}(\mathbb{E}[\mathbf{Y} \mid \mathbf{X}, i]) \\
&= \sum_{i=1}^K \pi_i \sigma_i^2 + \sum_{i=1}^K \pi_i \|\boldsymbol{\mu}_i - \bar{\boldsymbol{\mu}}\|^2.
\end{align}

Under the homogeneous mode variance assumption $\sigma_i^2 = \sigma^2$ for all $i$:
\begin{equation}
\mathrm{Var}(\mathbf{Y} \mid \mathbf{X}) = \sigma^2 + \sum_{i=1}^K \pi_i \|\boldsymbol{\mu}_i - \bar{\boldsymbol{\mu}}\|^2.
\end{equation}

\textit{Step 2 (Conditional Variance):} Conditioned on the correctly distilled reasoning embedding $\mathbf{e}_R$, the multimodal ambiguity collapses to the true single trajectory mode $k$, yielding $\mathbf{Y} \mid \mathbf{E}_{\text{fused}} \sim \mathcal{N}(\boldsymbol{\mu}_k, \sigma^2 \mathbf{I})$. Thus:
\begin{equation}
\mathrm{Var}(\mathbf{Y} \mid \mathbf{E}_{\text{fused}}) = \sigma^2.
\end{equation}

\textit{Step 3 (Variance Difference):} Subtracting the conditional variance from the unconditional mixture variance yields:
\begin{equation}
\begin{aligned}
&\mathrm{Var}(\mathbf{Y} \mid \mathbf{X}) - \mathrm{Var}(\mathbf{Y} \mid \mathbf{E}_{\text{fused}}) \\
&\qquad = \sum_{i=1}^K \pi_i \|\boldsymbol{\mu}_i - \bar{\boldsymbol{\mu}}\|^2 \ge 0.
\end{aligned}
\end{equation}

Because the between-mode variance $\sum_{i=1}^K \pi_i \|\boldsymbol{\mu}_i - \bar{\boldsymbol{\mu}}\|^2$ is strictly positive for non-identical mode centers, the reasoning prior strictly reduces predictive uncertainty. If within-mode variances differ ($\sigma_k^2 \neq \sigma_i^2$), the inequality holds whenever the variance of the true mode $\sigma_k^2$ does not exceed the aggregate mixture variance $\sum_{i=1}^K \pi_i \sigma_i^2 + \sum_{i=1}^K \pi_i \|\boldsymbol{\mu}_i - \bar{\boldsymbol{\mu}}\|^2$. \hfill $\square$

\subsection{The Bias-Variance Tradeoff in Joint Optimization}

While Proposition~\ref{prop:variance_reduction} demonstrates uncertainty reduction under successful mode selection, imperfect language generation could theoretically cause the Student LLM to isolate an incorrect mode $j \neq k$. 

\paragraph{Bias Formulation.} If the projected prior $\mathbf{e}_{R,j}$ directs the TSFM to sample from mode $j \sim \mathcal{N}(\boldsymbol{\mu}_j, \sigma_j^2 \mathbf{I})$ while the ground truth follows $k \sim \mathcal{N}(\boldsymbol{\mu}_k, \sigma_k^2 \mathbf{I})$, the conditional variance remains low ($\sigma_j^2$), but the prediction incurs structural bias:
\begin{equation}
\begin{aligned}
\mathrm{Bias}(\hat{\mathbf{Y}}) &= \mathbb{E}[\hat{\mathbf{Y}} \mid \mathbf{e}_{R,j}] - \mathbb{E}[\mathbf{Y} \mid \text{mode } k] \\
&= \boldsymbol{\mu}_j - \boldsymbol{\mu}_k.
\end{aligned}
\end{equation}

Evaluating the expected mean squared error under this misspecified distributional prediction gives:
\begin{equation}
\begin{aligned}
&\mathbb{E}\left[\|\mathbf{Y} - \hat{\mathbf{Y}}\|^2 \mid \mathbf{e}_{R,j}\right] = \underbrace{\|\boldsymbol{\mu}_j - \boldsymbol{\mu}_k\|^2}_{\mathrm{Bias}^2} \\
&\qquad + \underbrace{\mathrm{Tr}(\sigma_j^2 \mathbf{I})}_{\mathrm{Predictive\ Variance}} + \underbrace{\mathrm{Tr}(\sigma_k^2 \mathbf{I})}_{\mathrm{Irreducible\ Error}}.
\end{aligned}
\end{equation}

\paragraph{Bias Mitigation Mechanism.} When the hallucinated or misspecified mode $j$ diverges significantly from the ground truth trajectory $k$, the squared bias term $\|\boldsymbol{\mu}_j - \boldsymbol{\mu}_k\|^2$ dominates the empirical error. Under our joint objective:
\begin{equation}
\mathcal{L}_{\text{total}} = \alpha \mathcal{L}_{\text{CE}} + \beta \mathcal{L}_{\text{Quantile}},
\end{equation}
the quantile loss $\mathcal{L}_{\text{Quantile}}$ directly penalizes this distributional divergence. Large corrective gradients $\nabla_{\mathbf{E}_{\text{fused}}} \mathcal{L}_{\text{Quantile}}$ propagate backward through $\mathbf{W}_{\text{proj}}$ into the Student LLM parameters $\Theta_{\text{LLM}}$. Consequently, while $\mathcal{L}_{\text{CE}}$ drives variance reduction by encouraging mode isolation, joint backpropagation with $\mathcal{L}_{\text{Quantile}}$ provides a numerical anchor that suppresses mode misspecification and penalizes hallucinated trajectories.

\section{Extended Related Works}
\paragraph{Dataset-Specific Architectures.}
\label{app:extend_related}
Prior to the advent of large-scale pretraining, the state-of-the-art in time series forecasting relied on training specialized deep learning architectures from scratch on individual datasets. Models such as PatchTST \citep{nie2023a} introduced patching mechanisms to preserve local temporal semantics and reduce computational complexity. iTransformer \citep{liuitransformer} inverted the traditional attention mechanism to focus on variate correlations rather than temporal tokens, while TimeMachine \citep{ahamed2024timemachine} leveraged highly efficient architectures to capture long-term dependencies with linear complexity. While these models deliver strong performance on their specific target distributions, they lack zero-shot generalization capabilities. Furthermore, they operate entirely within the numerical domain, offering no textual interpretability or reasoning mechanisms to explain their strategies.

\paragraph{Explainable Forecasting.}
Traditional Explainable AI (XAI) methods for time series, such as SHAP~\citep{lundberg2017unified} or LIME~\citep{ribeiro2016should}, focus on feature attribution by identifying \textit{which} past time steps influenced a prediction \citep{rojat2021explainable}. While useful for debugging, these methods lack causal depth; they might highlight that time step $t$ was important, but cannot explain \textit{why} (e.g., a seasonal shift or fiscal event). By generating explicit natural language reasoning traces aligned with cross-channel dependencies, our approach achieves ``inherent interpretability'' \citep{rudin2019stop}. By evaluating against rigorous reasoning benchmarks like TFRBench, we ensure our framework provides users with a transparent narrative that can be inspected alongside the forecast.

\section{Algorithm for Training and Inference}
\label{app:algorithms} 
\begin{algorithm}
\small
\caption{Reasoning-Aware Joint Training for LLM and TSFM}
\label{alg:method}
\begin{algorithmic}[1]
\REQUIRE Training time series dataset $\mathcal{D}$, Large LLM, Small-LLM parameterized by $\Theta_{LLM}$, Target TSFM, Projection matrix $\mathbf{W}_{proj}$, Loss weights $\alpha, \beta \in [0, 1]$.

\vspace{0.2cm}
\STATE \textbf{Phase 1: Reference and Baseline Reasoning Generation (Offline)}
\FOR{each sampled window $(\mathbf{X}, \mathbf{Y}) \in \mathcal{D}$}
    \STATE Retrieve time-bound external events $\mathcal{E}$ corresponding to the temporal window.
    \STATE $\mathbf{R}_{ref} \leftarrow \text{ReferenceLLM}(\mathbf{X}, \mathbf{Y}, \mathcal{E})$ \COMMENT{Generate concrete reference reasoning via TFRBench}
    \STATE $\mathbf{R}_{base} \leftarrow \text{LargeLLM}(\mathbf{X})$ \COMMENT{Generate baseline reasoning from historical context}
\ENDFOR

\vspace{0.2cm}
\STATE \textbf{Phase 2: Joint Optimization Framework (Online)}
\WHILE{not converged}
    \STATE Sample a batch of $(\mathbf{X}, \mathbf{Y}, \mathbf{R}_{ref}, \mathbf{R}_{base})$.
    
    \vspace{0.2cm}
    \STATE \COMMENT{\textcolor{blue}{\textit{Step 1: Reasoning Distillation}}}
    \STATE $P_{in} \leftarrow \text{FormatPrompt}(\mathbf{X}, \text{metadata}, \mathbf{R}_{base})$ 
    \STATE $\mathbf{H} \leftarrow \text{SmallLLM}_{hidden\_states}(P_{in})$ \COMMENT{Extract last hidden states $\in \mathbb{R}^{L \times d_{llm}}$}
    \STATE Compute $\mathcal{L}_{CE} = - \sum_{j=1}^{M} \log P(\mathbf{R}_{ref, j} \mid P_{in}, \mathbf{R}_{ref, <j} ; \Theta_{LLM})$
    
    \vspace{0.2cm}
    \STATE \COMMENT{\textcolor{blue}{\textit{Step 2: Cross-Modal Latent Projection}}}
    \STATE $\mathbf{h}_{R} \leftarrow \frac{1}{L} \sum_{i=1}^{L} \mathbf{H}_{i}$ \COMMENT{Mean-pool over the sequence dimension}
    \STATE $\mathbf{e}_{R} \leftarrow \mathbf{h}_{R} \mathbf{W}_{proj}$ \COMMENT{Project reasoning into TSFM embedding space $\in \mathbb{R}^{d_{ts}}$}
    
    \vspace{0.2cm}
    \STATE \COMMENT{\textcolor{blue}{\textit{Step 3: Latent Fusion and Quantile Forecasting}}}
    \STATE $\mathbf{E}_{TS} \leftarrow \text{TSFM\_Embed}(\mathbf{X})$ \COMMENT{Extract native time series sequence embeddings}
    \STATE $\mathbf{E}_{fused} \leftarrow \mathbf{e}_{R} \oplus \mathbf{E}_{TS}$ \COMMENT{Fuse via generalized operator and duplicate across variates}
    \STATE $\hat{\mathbf{Y}}_q \leftarrow \text{TSFM\_Decode}(\mathbf{E}_{fused})$ \COMMENT{Predict future quantiles for levels $\mathcal{Q}$}
    \STATE Compute $\mathcal{L}_{Quantile}$ between predictions $\hat{\mathbf{Y}}_q$ and ground truth $\mathbf{Y}$
    
    \vspace{0.2cm}
    \STATE \COMMENT{\textcolor{blue}{\textit{Step 4: End-to-End Backpropagation}}}
    \STATE $\mathcal{L}_{total} \leftarrow \alpha \mathcal{L}_{CE} + \beta \mathcal{L}_{Quantile}$
    \STATE Update $\mathbf{W}_{proj}$, TSFM parameters, and $\Theta_{LLM}$ (via LoRA) using $\nabla \mathcal{L}_{total}$
\ENDWHILE
\end{algorithmic}
\end{algorithm}

% \section{Algorithm for Inference Phase}
% \label{app:algo_inference}
\begin{algorithm}[H]
\small
\caption{Dual-Generation Inference Phase (Evaluation)}
\label{alg:inference}
\begin{algorithmic}[1]
\REQUIRE Unseen historical time series $\mathbf{X} \in \mathbb{R}^{T \times V}$, Large LLM, Small-LLM parameterized by $\Theta_{LLM}$, Target TSFM, Projection matrix $\mathbf{W}_{proj}$.

\vspace{0.2cm}
\STATE \COMMENT{\textcolor{blue}{\textit{Step 1: Baseline Context Initialization}}}
\STATE $\mathbf{R}_{base} \leftarrow \text{LargeLLM}(\mathbf{X})$ \COMMENT{Generate baseline reasoning from historical context}
\STATE $P_{in} \leftarrow \text{FormatPrompt}(\mathbf{X}, \text{metadata}, \mathbf{R}_{base})$

\vspace{0.2cm}
\STATE \COMMENT{\textcolor{blue}{\textit{Step 2: Strategic Directive Generation}}}
\STATE $\hat{\mathbf{R}}, \mathbf{H} \leftarrow \text{SmallLLM}(P_{in})$ \COMMENT{Autoregressively generate directive for future prediction and extract hidden states $\in \mathbb{R}^{L \times d_{llm}}$}

\vspace{0.2cm}
\STATE \COMMENT{\textcolor{blue}{\textit{Step 3: Cross-Modal Latent Projection}}}
\STATE $\mathbf{h}_{R} \leftarrow \frac{1}{L} \sum_{i=1}^{L} \mathbf{H}_{i}$ \COMMENT{Mean-pool over the sequence dimension}
\STATE $\mathbf{e}_{R} \leftarrow \mathbf{h}_{R} \mathbf{W}_{proj}$ \COMMENT{Project reasoning into TSFM embedding space $\in \mathbb{R}^{d_{ts}}$}

\vspace{0.2cm}
\STATE \COMMENT{\textcolor{blue}{\textit{Step 4: Latent Fusion and Numerical Forecasting}}}
\STATE $\mathbf{E}_{TS} \leftarrow \text{TSFM\_Embed}(\mathbf{X})$ \COMMENT{Extract native time series sequence embeddings}
\STATE $\mathbf{E}_{fused} \leftarrow \mathbf{e}_{R} \oplus \mathbf{E}_{TS}$ \COMMENT{Fuse via sequence prefixing or initial state substitution}
\STATE $\hat{\mathbf{Y}}_q \leftarrow \text{TSFM\_Decode}(\mathbf{E}_{fused})$ \COMMENT{Predict future quantiles for levels $\mathcal{Q}$}

\vspace{0.2cm}
\RETURN $\hat{\mathbf{R}}, \hat{\mathbf{Y}}_q$ \COMMENT{Output both the explicit analytical strategy and continuous numerical forecasts}
\end{algorithmic}
\end{algorithm}
\section{Baseline Comparison}
\label{app:baseline}

\paragraph{Statistical Methods.} We include traditional baselines such as ARIMA~\citep{box2015time}, Seasonal Naive~\citep{aksu2025gifteval}, AutoARIMA~\citep{hyndman2008automatic}, AutoTheta~\citep{spiliotis2020generalizing}, and AutoETS following~\citep{liu2026timer}.

\paragraph{Time Series Foundation Models (TSFMs).} We benchmark against leading numerical foundation models including TimesFM-2.5~\citep{das2024decoder}, Chronos-2.0~\citep{ansari2025chronos}, Timer-S1~\citep{liu2026timer}, Migas-1.0~\citep{synthefy_migas_2025}, TiRex~\citep{auer2026tirex}, Moirai2~\citep{liu2025moirai}, Toto-1.0~\citep{cohen2024toto}, TabPFN-TS~\citep{hoo2025tables}, and Sundial-Base~\citep{liu2025sundial}. PatchTST~\citep{nie2023a} is related but is not a matched GIFT-Eval baseline, as it is generally trained and evaluated per task rather than as an adapter on a pretrained TSFM.

\paragraph{General-Purpose LLMs.} To demonstrate the modality gap inherent in zero-shot prompting, we evaluate against several models from the Gemini~\citep{comanici2025gemini, team2023gemini} and Claude families: Gemini-2.5-Flash, Gemini-2.5-Pro, Gemini-3.1-Pro, Claude-Sonnet-4, and Claude-Sonnet-4.5 following~\citep{ahamed2026tfrbench}. We used Vertex AI for these models~\footnote{\url{https://console.cloud.google.com/agent-platform/model-garden}}.

\paragraph{Task-Specific TS-LLMs.} We additionally compare text-to-number TS-LLM methods (Time-LLM~\citep{jin2024timellm}, PromptCast~\citep{xue2023promptcast}, and Chow-style CoT~\citep{chow2024towards}) on a multi-domain GIFT-Eval subset in Sec.~\ref{sec:ablation} and App.~\ref{app:tsllm_baselines}. These methods are trained or evaluated per task and are not matched pretrained-TSFM adapters under the full GIFT-Eval protocol.

\section{Additional Ablations and Experiments}
\label{app:additional_ablations}

This section presents additional ablation studies and experiments that isolate the contribution of reasoning-aware training beyond numerical fine-tuning and generic auxiliary embeddings, together with analysis on TFRBench and TS-LLM based baselines. Subsections~\ref{app:chronos_ft_control}--\ref{app:tsllm_baselines} covers the expanded ablations and experiments.

\subsection{No-Reasoning TSFM Fine-Tune}
\label{app:chronos_ft_control}
We fine-tune Chronos-2.0 on the same time-series windows used by \ours{}, but without a student LLM, projection module, or reasoning traces. From five GIFT-Eval domains we randomly select five datasets with distinct sampling frequencies and report MAE and MASE for Chronos-2.0, a finetuned version of Chronos-2.0 (Chronos-2 FT), and \ours{} (+Chronos-2). As shown in Table~\ref{tab:chronos_ft_control}, fine-tuning Chronos-2.0 alone typically improves over the Chronos-2.0 baseline, yet remains worse than \ours{}, indicating that numerical adaptation accounts for only part of the improvement while the remaining gap is attributable to reasoning-aware training.

\subsection{Time Series and Reasoning Shuffling}
\label{app:train_pairing_shuffle}
To test whether series--reasoning alignment matters during adaptation, we compare \ours{} (paired series--reasoning) to a shuffled regime in which reasoning texts are deranged across samples. Table~\ref{tab:train_pairing_shuffle} (main text) reports the performance. Shuffled scenario has higher error than \ours{}, indicating that preserving correspondence improves accuracy beyond merely exposing the model to reasoning-shaped text.

\subsection{Generic Summary Embedding}
\label{app:generic_summary_control}
We replace student reasoning hidden states with generic summary embeddings: for each dataset we use a short, non-forecasting descriptor of the series alone (e.g., domain and sampling frequency, with no strategy or predictive plan), map it to a vector in the same latent width as the student hidden state, and apply the same projection $W_{\mathrm{proj}}$ and Chronos-2 incorporation as in \ours{}. Table~\ref{tab:generic_summary_control} compares this generic-summary prior to \ours{} on five GIFT-Eval datasets. The matched-dimension generic channel does not recover the gains of \ours{}, especially where reasoning yields large improvements (e.g., Solar), indicating that the benefit of \ours{} is not explained by a generic extra vector of the same shape.

\begin{table}[H]
\centering
\footnotesize
\caption{Generic summary embedding on GIFT-Eval datasets with varying frequencies. We report MAE and MASE (lower is better). \textbf{Bold}: best; \underline{underline}: second best.}
\label{tab:generic_summary_control}
% \setlength{\tabcolsep}{3pt}
% \renewcommand{\arraystretch}{1.05}
% \resizebox{\linewidth}{!}{%
\begin{tabular}{lll *{2}{c} *{2}{c}}
\toprule
\multirow{2}{*}{\textbf{Dataset}} & \multirow{2}{*}{\textbf{Domain}} & \multirow{2}{*}{\textbf{Freq.}} &
\multicolumn{2}{c}{\textbf{Generic Summary}} &
\multicolumn{2}{c}{\textbf{\ours{}}} \\
\cmidrule(lr){4-5}\cmidrule(lr){6-7}
& & &
\textbf{MAE} & \textbf{MASE} &
\textbf{MAE} & \textbf{MASE} \\
\midrule
Solar & Energy & W &
$\underline{1.163\!\times\!10^{3}}$ & $\underline{1.270}$ &
$\mathbf{2.955\!\times\!10^{2}}$ & $\mathbf{0.315}$ \\
BizITObs L2C & CloudOps & 5T &
$\underline{5.029\!\times\!10^{0}}$ & $\underline{0.506}$ &
$\mathbf{4.321\!\times\!10^{0}}$ & $\mathbf{0.429}$ \\
KDD Cup 2018 & Nature & H &
$\underline{2.578\!\times\!10^{1}}$ & $\underline{1.093}$ &
$\mathbf{1.662\!\times\!10^{1}}$ & $\mathbf{0.721}$ \\
US Births & Healthcare & M &
$\mathbf{4.641\!\times\!10^{3}}$ & $\mathbf{0.525}$ &
$\underline{5.532\!\times\!10^{3}}$ & $\underline{0.627}$ \\
Hierarchical Sales & Sales & D &
$\underline{2.391\!\times\!10^{0}}$ & $\underline{0.765}$ &
$\mathbf{2.273\!\times\!10^{0}}$ & $\mathbf{0.738}$ \\
\bottomrule
\end{tabular}
% }
\end{table}

\subsection{Uncertainty of Reported Metrics}
\label{app:metric_uncertainty}
\begin{figure}[H]
\centering
\includegraphics[width=\linewidth]{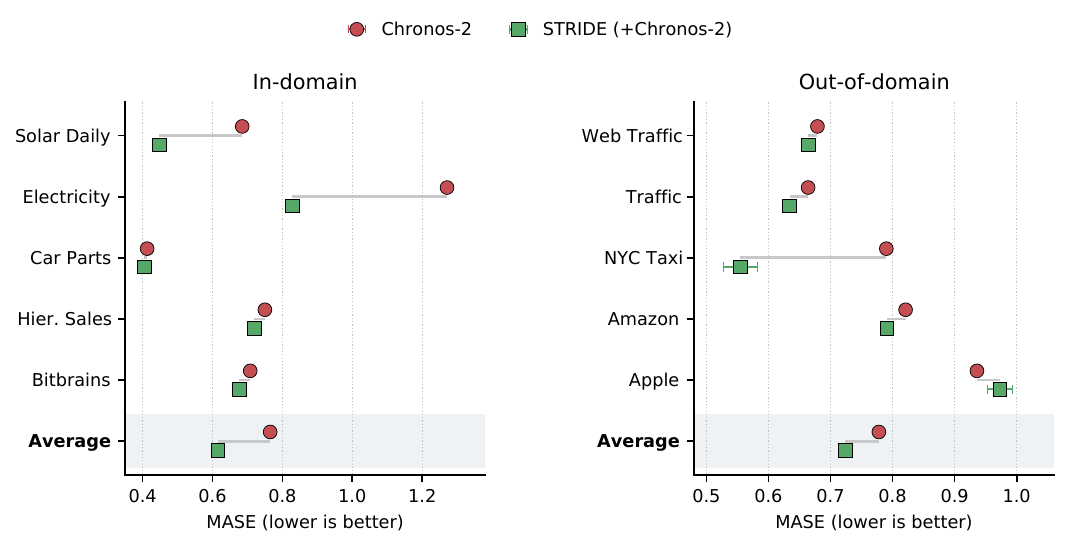}
\caption{TFRBench in-domain and OOD MASE for Chronos-2 and \ours{} (+Chronos-2), from Tables~\ref{tab:numerical_tfrbench_indomain} and~\ref{tab:numerical_tfrbench_ood}. Markers are evaluation-seed means; whiskers are the reported standard deviation (lower is better). Highlighted rows are averages.}
\label{fig:tfrbench_mase_uncertainty}
\end{figure}

Tables~\ref{tab:numerical_tfrbench_indomain} and~\ref{tab:numerical_tfrbench_ood} report TFRBench MAE and MASE as $mean_{std}$ over evaluation seeds. Figure~\ref{fig:tfrbench_mase_uncertainty} compares Chronos-2 and \ours{} (+Chronos-2) MASE on the in-domain and out-of-domain (OOD) splits. \ours{} attains a lower mean MASE than Chronos-2 on all five in-domain datasets and on four of five OOD datasets. Split averages are $0.615_{0.010}$ vs.\ $0.765_{0.000}$ in-domain and $0.724_{0.011}$ vs.\ $0.778_{0.000}$ OOD. Chronos-2 uses deterministic decoding, so its reported MASE standard deviation is $0$. As a conservative check, we add one reported standard deviation to each \ours{} mean. \ours{} remains lower-error on every dataset where the means already favored \ours{}, including the split averages ($0.625$ vs.\ $0.765$ in-domain; $0.735$ vs.\ $0.778$ OOD).

\subsection{Human Validation of Reasoning Evaluation}
\label{app:human_eval_tfrbench}
Numerical forecasting benchmarks such as GIFT-Eval evaluate point and distributional error (MAE, MASE, CRPS) but do not score the reasoning that accompanies a forecast. To the best of our knowledge, TFRBench~\citep{ahamed2026tfrbench} is currently the latest benchmark dedicated to forecasting reasoning evaluation. It scores generated narratives on a 1--5 rubric covering Domain Relevance, Forecasting Correctness, Event Relevance, and Logic-to-Number Consistency. We adopt this protocol to assess reasoning quality.

Besides, TFRBench calibrates its LLM-as-a-Judge against a blinded human expert panel, reporting a Pearson correlation of 70.4\% between the human ensemble and the automated verifier, and strong rank agreement among independent LLM judges (Spearman $0.842$).

\subsection{Comparison with TS-LLM Baselines}
\label{app:tsllm_baselines}
We compare \ours{} to relevant task-specific TS-LLM methods (Time-LLM~\citep{jin2024timellm}, PromptCast~\citep{xue2023promptcast}, and Chow-style CoT~\citep{chow2024towards}) on the five-domain GIFT-Eval datasets. Time-LLM keeps the LLM backbone frozen, reprograms patched series into text prototypes, and projects the resulting hidden states to a numeric forecast. PromptCast treats forecasting as sentence-to-sentence generation over serialized history. ~\citet{chow2024towards} fine-tune an LLM on chain-of-thought time-series reasoning rather than token-level numeric decoding; our Chow-style baseline follows that CoT idea by requesting a reasoning chain before emitting numeric forecast tokens. Table~\ref{tab:tsllm_baselines} shows that all three remain higher-error than \ours{} in both MAE and MASE on the datasets.

\begin{figure*}
    \centering
    \includegraphics[width=\linewidth]{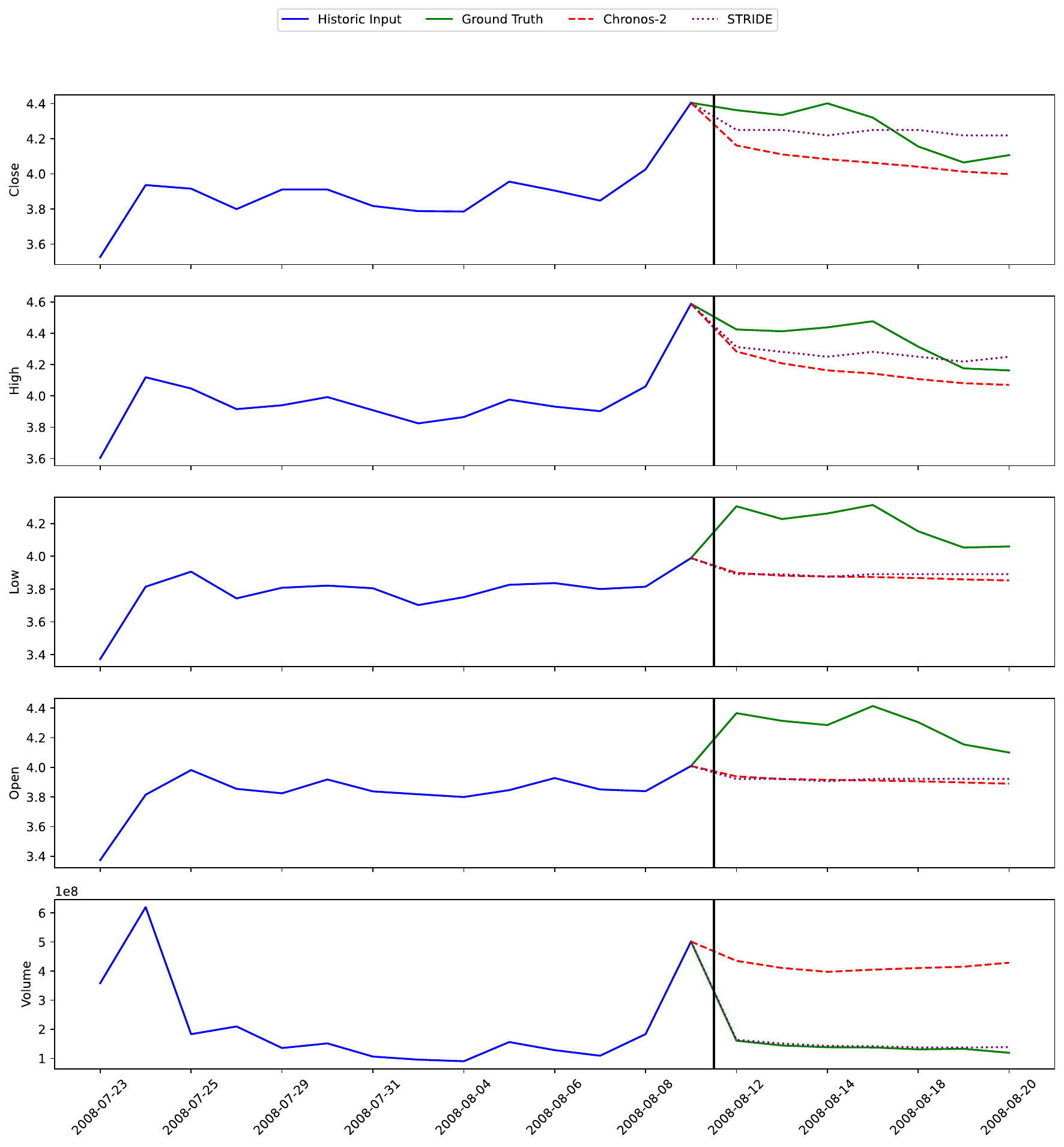}
    \caption{Forecasting prediction for Amazon dataset.}
    \label{fig:amazon_case}
\end{figure*}

\section{Implementation Details}
\label{app:implementation}

For our knowledge distillation pipeline, primarily, we utilize Gemini-3.1-Pro as the Teacher LLM to generate the high-quality reference reasoning utilizing TFRBench~\citep{ahamed2026tfrbench}. For baseline reasoning provider model also, we used Gemini-3.1-pro. The lightweight student model (Small-LLM) is primarily parameterized by Gemma-3-4B-it~\citep{team2025gemma}. To maintain training stability and preserve fundamental linguistic capabilities, we apply Low-Rank Adaptation (LoRA)~\citep{hu2021lora} to the attention mechanisms of the Small-LLM during the joint optimization phase. Detailed prompts for these LLMs are provided in App.~\ref{app:prompts_llm}. The entire pipeline is trained end-to-end using a combination of textual cross-entropy loss and numerical quantile loss, while providing equal weights ($\alpha,\beta=1$) to both terms in Eq.~\ref{eq:total_loss}. For optimization we used AdamW optimizer~\citep{loshchilov2018decoupled}. To train our framework, we utilized a distributed computing environment configured with either 16 NVIDIA A100 (40GB) for Chronos-2.0 or 8 NVIDIA H100 (80GB) GPUs for Timer-S1.
Besides Gemini-3.1-pro and Gemma-3-4B-it, we also demonstrate the effectiveness of \ours{} with Qwen3-4B-Instruct~\citep{yang2025qwen3} and Claude-Sonnet-4.5~\citep{anthropic2025sonnet45} in \textsection~\ref{subsec:small_llm}.
While we adopt Chronos-2.0 as TSFM backbone, we demonstrate generalizability of our framework with SOTA Timer-S1.

Chronos-2.0 is a $\approx$120M-parameter TSFM, while Timer-S1 is an $\approx$8.3B-parameter mixture-of-experts TSFM; the student LLMs have about 4B parameters, of which only the LoRA adapters (rank 16) are trained, jointly with the projection layer $\mathbf{W}_{proj}$ and the TSFM. We follow the default training recipes of the respective backbones and fixed all hyperparameters a priori without tuning on any evaluation data; a full training run takes roughly 24 hours. All prompts and reasoning traces are in English. Forecasting metrics are computed with the official GIFT-Eval evaluation code over its full suite of 97 tasks and with the TFRBench evaluation scripts on 3 evaluation runs, and reasoning quality is scored with the TFRBench LLM-as-a-Judge protocol (prompt in App.~\ref{app:prompts_llm}).

\section{Forecasting Case Studies}
\label{app:forecasting_case}
In this section, we demonstrate several case studies, where we compare \ours{} framework with original Chronos-2 method.

\begin{figure}
    \centering
    \includegraphics[width=\linewidth]{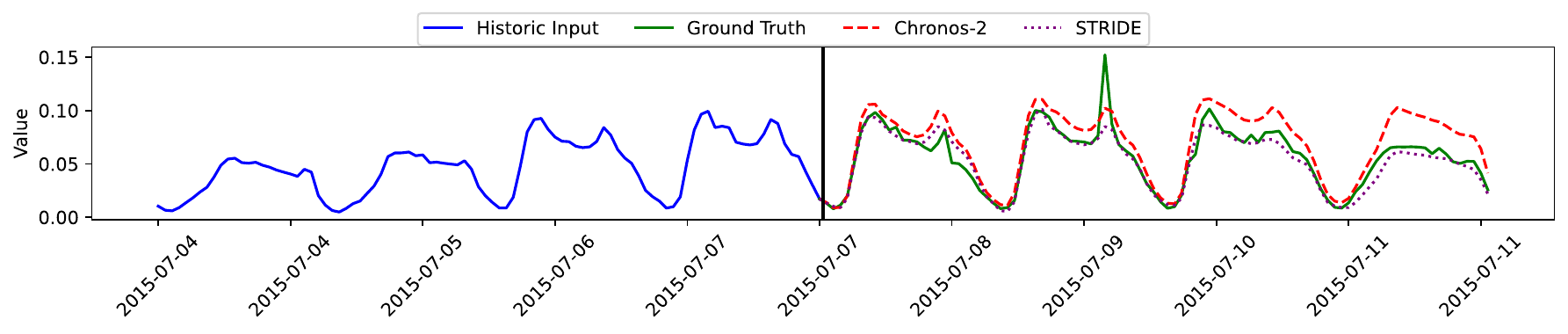}
    \caption{Forecasting prediction for Traffic dataset.}
    \label{fig:traffic_case}
\end{figure}
\begin{figure}
    \centering
    \includegraphics[width=\linewidth]{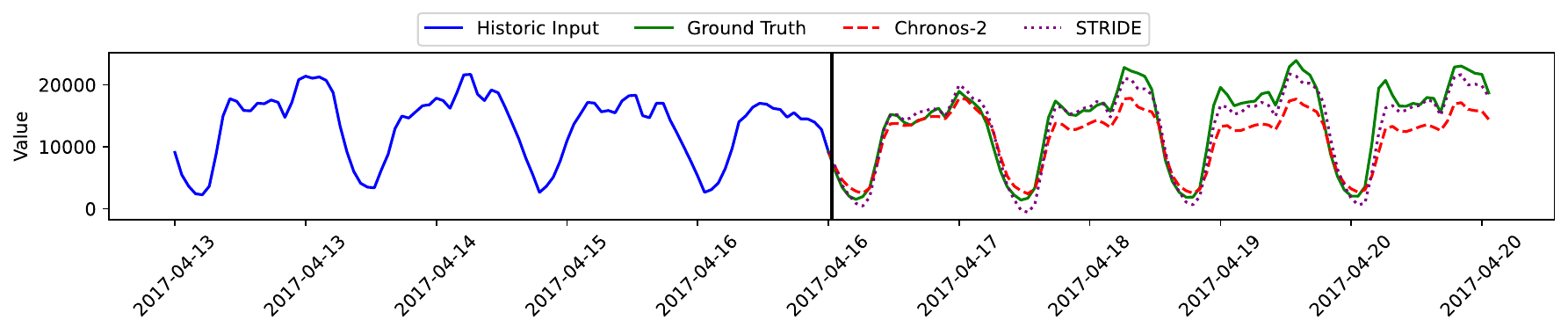}
    \caption{Forecasting prediction for Nyc Taxi dataset.}
    \label{fig:nyc_taxi_case}
\end{figure}

\section{Reasoning Case Studies}
\label{app:reasoning_case}

\begin{tcolorbox}[
  colback=blue!3!white,
  colframe=blue!60!black,
  title={\textbf{NYC Taxi}},
  fonttitle=\bfseries,
  boxsep=2mm,
  arc=2mm,
  breakable % This option allows the box to flow onto the next page
]

% --- Figure Placeholder ---
\begin{center}
    % Replace 'example-image' with your actual image path, e.g., 'figures/nyc_taxi_plot.pdf'
    \includegraphics[width=0.9\textwidth]{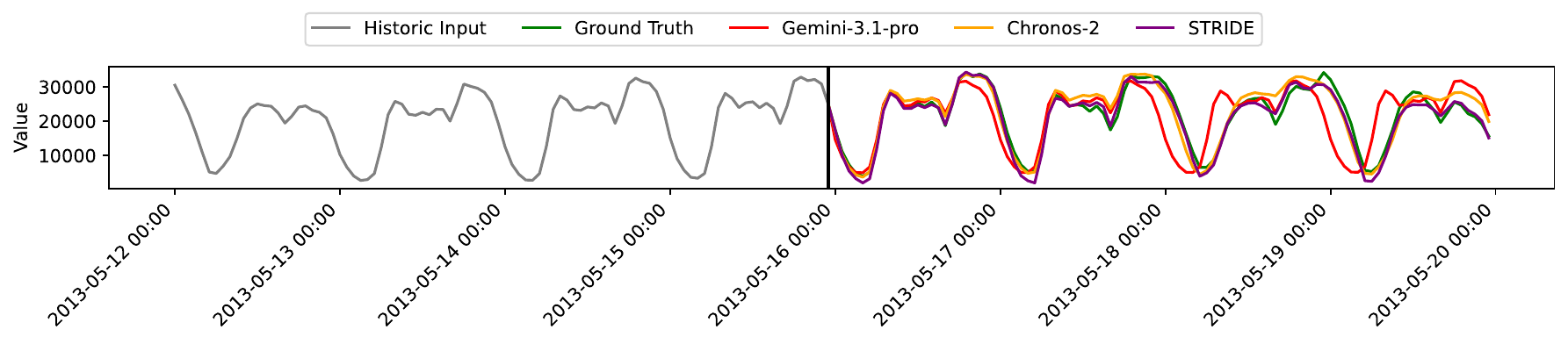} 
    \vspace{2mm} \\
    % \textit{Figure: NYC Taxi forecasting comparison between Baseline (Gemini-3.1-pro) and Ours.}
\end{center}
\vspace{2mm}
\hrule
\vspace{3mm}
% --------------------------

\textbf{Baseline Reasoning (Gemini-3.1-pro)} \\
The time series exhibits a strong 24-hour seasonality, typical of hourly taxi trip counts, with deep troughs in the early morning (03:00-04:00) and peaks in the evening (18:00-19:00). The trend component shows a steady increase over the historical period, rising from approximately 18,000 to over 21,000. To forecast the next 96 steps (4 days), we project the trend to continue its gradual upward trajectory, starting at a baseline of 21,500 and increasing by 2 units per hour. We then superimpose the extracted 24-hour seasonal profile onto this projected trend. Residuals are assumed to be zero for the forecast horizon. This additive combination of a linearly increasing trend and a fixed daily seasonal pattern yields the final predicted values, maintaining the expected order of magnitude and daily structural variations.

\vspace{3mm}
\hrule
\vspace{3mm}

\textbf{\ours{}'s Reasoning}
\begin{itemize}
    \setlength{\itemsep}{1pt}
    \item \textbf{Cross-Channel Analysis:} As a single-channel dataset, NYC taxi trip volume is intrinsically driven by the city's 24-hour mobility cycle, commuter schedules, and nightlife. The data exhibits a strong dependency on the day of the week, with distinct behavioral shifts between weekdays and weekends dictating the flow of urban transit.
    \item \textbf{Trend Component Plan:} The overarching baseline hovers around a mean of approximately 20,000 trips per hour, but this is heavily modulated by the day of the week. Weekdays sustain higher overall daytime activity, while the weekend trend shifts the bulk of the volume toward the afternoon and late evening, reflecting a transition from business-driven to leisure-driven demand.
    \item \textbf{Seasonality Component Plan:} The seasonal structure is defined by a robust 24-hour cycle that morphs based on the day. On weekdays (first 48 hours), the pattern is distinctly bimodal: a sharp morning commuter peak around 8:00 AM reaching 26,000-28,000 trips, a midday plateau near 24,000 trips, and a dominant evening rush peaking between 7:00 PM and 8:00 PM at 33,000-34,000 trips. On weekends (latter 48 hours), the morning commute peak vanishes. Instead, demand builds steadily throughout the afternoon. Saturday maintains a strong evening peak of around 31,000 trips, while Sunday's peak is softer and earlier. Across all days, an early morning lull occurs between 3:00 AM and 5:00 AM.
    \item \textbf{Residuals \& Event Plan:} Specific structural anomalies are embedded in the daily flow. Weekdays feature a sharp, temporary dip around 4:00 PM (dropping to roughly 18,000-19,000 trips), perfectly aligning with the traditional NYC taxi driver shift change. The transition into the weekend introduces a "nightlife buffer"; late-night hours on Friday and Saturday (midnight to 2:00 AM) sustain unusually high volumes of 21,000-29,000 trips. The model must anticipate these localized structural deviations rather than smoothing them out.
\end{itemize}

\vspace{3mm}
\hrule
\vspace{3mm}

\textbf{Evaluation Scores} \\
\textbf{Overall Scores:} \ours{}: 4.0 \quad|\quad Gemini-3.1-pro: 2.75

\vspace{2mm}

\vspace{2mm}
\textbf{Metrics Breakdown (Gemini-3.1-pro):}
\begin{itemize}
    \setlength{\itemsep}{1pt}
    \item \textbf{Domain Relevance (3/5):} Uses basic terminology but lacks true domain expert depth, failing to distinguish weekday vs. weekend patterns.
    \item \textbf{Forecasting Correctness (2/5):} Identifies basic intra-day seasonality but misses weekly seasonality and incorrectly models a continuous linear upward trend.
    \item \textbf{Event Relevance (1/5):} Completely ignores all external events, explicitly stating residuals are assumed to be zero.
    \item \textbf{Logic Consistency (5/5):} Perfectly aligns with the candidate's stated reasoning, but the methodology itself is flawed.
\end{itemize}
\textbf{Metrics Breakdown (\ours{}):}
\begin{itemize}
    \setlength{\itemsep}{1pt}
    \item \textbf{Domain Relevance (5/5):} Demonstrates expert-level domain knowledge, using precise terminology such as '24-hour mobility cycle,' 'bimodal,' 'taxi driver shift change,' and 'nightlife buffer'.
    \item \textbf{Forecasting Correctness (4/5):} Correctly identifies the overarching trend and the complex daily/weekly seasonality. Misses specific magnitude spikes caused by localized external events.
    \item \textbf{Event Relevance (2/5):} Misses the specific external events (sports games, festivals) highlighted in the Ground Truth, despite effectively capturing recurring structural events.
    \item \textbf{Logic Consistency (5/5):} The numerical forecast is a perfect translation of the narrative. Specific claims are precisely reflected in the generated numbers.
\end{itemize}

\end{tcolorbox}

%%%%%%%%%%%%%%%%%%%%%%%%%%%%%%%%%%%%%%%%%%%%%%

\begin{tcolorbox}[
  colback=blue!3!white,
  colframe=blue!60!black,
  title={\textbf{Traffic}},
  fonttitle=\bfseries,
  boxsep=2mm,
  arc=2mm,
  breakable % This option allows the box to flow onto the next page
]

% --- Figure Placeholder ---
\begin{center}
    % Replace 'example-image' with your actual image path, e.g., 'figures/traffic_plot.pdf'
    \includegraphics[width=0.9\textwidth]{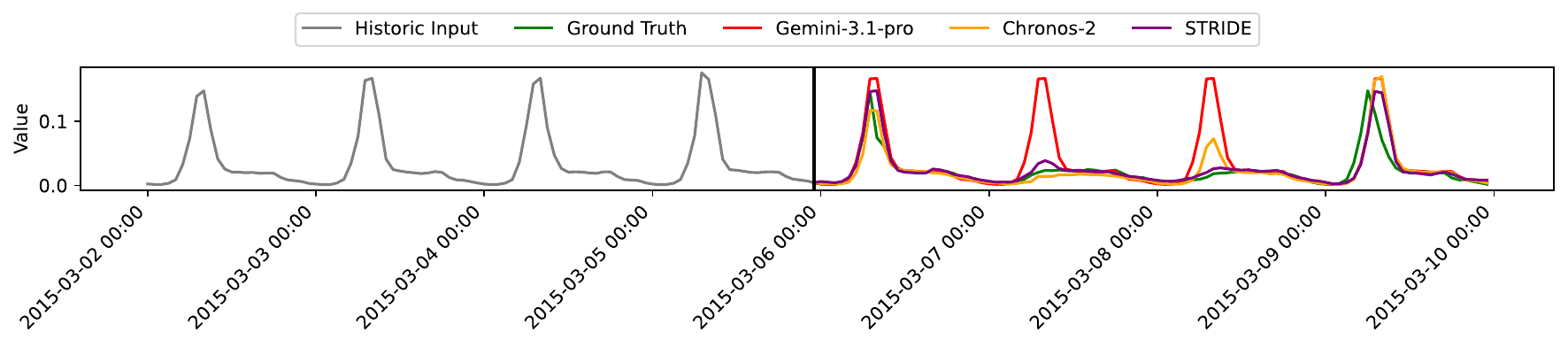} 
    \vspace{2mm} \\
    % \textit{Figure: Traffic forecasting comparison between Baseline (Gemini-3.1-pro) and Ours.}
\end{center}
\vspace{2mm}
\hrule
\vspace{3mm}
% --------------------------

\textbf{Baseline Reasoning (Gemini-3.1-pro)} \\
The time series exhibits a strong and consistent daily seasonality (24-hour period) with a relatively stable trend. By analyzing the pre-computed decomposition features, we observe that the trend component stabilizes around 0.0348 towards the end of the historical data. The seasonal component repeats perfectly every 24 hours, capturing the daily peaks (around hours 07:00 and 08:00) and troughs (around hours 00:00 to 02:00). To forecast the next 96 steps (4 days), we project the constant trend of 0.0348 forward and add it to the corresponding hour's seasonal component. This approach generates a repeating 24-hour pattern that closely matches the magnitude and shape of the historical data, ensuring all predicted values remain within the expected 0.00 to 0.18 range.

\vspace{3mm}
\hrule
\vspace{3mm}

\textbf{\ours{}'s Reasoning}
\begin{itemize}
    \item \textbf{Cross-Channel Analysis:} As this is a single-channel dataset, the analysis focuses exclusively on the internal temporal dynamics of the traffic volume. The data is inherently cyclical, driven by human mobility patterns, and exhibits strong daily and weekly periodicities without reliance on external covariates.
    
    \item \textbf{Trend Component Plan:} The underlying baseline trend remains relatively flat and stable throughout the forecast horizon, hovering near a low baseline of approximately 0.002 to 0.01 units during off-peak nighttime hours. Rather than a continuous upward or downward drift, the trend acts as a stationary foundation that dynamically scales to accommodate the varying magnitudes of daily traffic surges.
    
    \item \textbf{Seasonality Component Plan:} The traffic volume is dominated by a strict 24-hour seasonal cycle, featuring pronounced peaks during specific active hours (roughly the 7th to 9th hour of the daily cycle). Crucially, this daily seasonality is modulated by a broader weekly pattern. On high-activity days (e.g., typical weekdays), the seasonal amplitude surges to roughly 0.14 to 0.15 units. On low-activity days (e.g., weekends), this amplitude is significantly dampened, reaching only 0.025 to 0.04 units. The model must accurately capture both the timing of the daily peaks and these day-to-day amplitude shifts.
    
    \item \textbf{Residuals \& Predicted Event Plan:} The residual component should remain tightly constrained, representing minor, random fluctuations in traffic flow with deviations of roughly 0.001 to 0.005 units. The most notable "event" in this sequence is the structural shift during the middle 48 hours of the forecast, where peak traffic volume drops drastically by over 70\% (from $\sim$0.14 down to $\sim$0.03 units). This reflects a predictable weekend or holiday effect. The model must anticipate this temporary suppression of traffic and subsequently predict a full recovery to the baseline high-volume pattern (returning to $\sim$0.14 units) by the final day of the forecast window.
\end{itemize}

\vspace{3mm}
\hrule
\vspace{3mm}

\textbf{Evaluation Scores} \\
\textbf{Overall Scores:} \ours{}: 4.75 \quad|\quad Gemini-3.1-pro: 3.0

\vspace{2mm}

\textbf{Metrics Breakdown (Gemini-3.1-pro):}
\begin{itemize}
    \setlength{\itemsep}{1pt}
    \item \textbf{Domain Relevance (2/5):} Uses generic time series terminology (e.g., 'trend', 'seasonal component', '24-hour period') but fails to incorporate any domain-specific language related to traffic, such as 'commute hours', 'weekdays vs. weekends', or 'congestion'.
    \item \textbf{Forecasting Correctness (2/5):} Correctly identifies the daily seasonality but completely misses the crucial weekly seasonality (the difference between weekdays and weekends). It erroneously predicts sharp morning commute peaks for Saturday and Sunday.
    \item \textbf{Event Relevance (3/5):} No external events were mentioned by the candidate. While it missed the opportunity to discuss the lack of events or historical weather impacts, it did not hallucinate any false events, making it acceptable but lacking depth.
    \item \textbf{Logic Consistency (5/5):} The numerical forecast perfectly aligns with the candidate's stated logic. The reasoning explicitly states it will generate a repeating 24-hour pattern, and the predicted values reflect exactly the same 24-hour cycle for all four days.
\end{itemize}

\vspace{2mm}

\textbf{Metrics Breakdown (\ours{}):}
\begin{itemize}
    \setlength{\itemsep}{1pt}
    \item \textbf{Domain Relevance (5/5):} Demonstrates expert domain knowledge, correctly identifying the dataset as traffic volume/occupancy and using precise terminology such as 'commuter traffic patterns,' 'intra-day and intra-week cyclicality,' and 'morning/evening commute peaks.'
    \item \textbf{Forecasting Correctness (4/5):} Correctly identifies the overall trend and the stark contrast between weekday and weekend seasonality. However, it slightly misses the exact shape of the weekend traffic, predicting a suppressed morning peak rather than the broad afternoon plateau described in the Ground Truth.
    \item \textbf{Event Relevance (5/5):} Does not hallucinate any external events and correctly identifies the transition to the weekend as the primary structural shift impacting the forecast window, which aligns perfectly with the Ground Truth's lack of external events for those days.
    \item \textbf{Logic Consistency (5/5):} The numerical forecast is a precise translation of the reasoning. The text explicitly predicts weekday peaks of approx. 0.147 and weekend peaks of approx. 0.038 and 0.027, which are exactly reflected in the generated data points.
\end{itemize}

\end{tcolorbox}

%%%%%%%%%%%%%%%%%%%%%%%%%%%%%%%%%%%%%%%%%%%%%%%%%%%%%%%%

\begin{tcolorbox}[
  colback=blue!3!white,
  colframe=blue!60!black,
  title={\textbf{Amazon Pricing}},
  fonttitle=\bfseries,
  boxsep=2mm,
  arc=2mm,
  breakable % This option allows the box to flow onto the next page
]

% --- Figure Placeholder ---
\begin{center}
    % Replace 'example-image' with your actual image path, e.g., 'figures/amazon_plot.pdf'
    \includegraphics[width=0.9\textwidth]{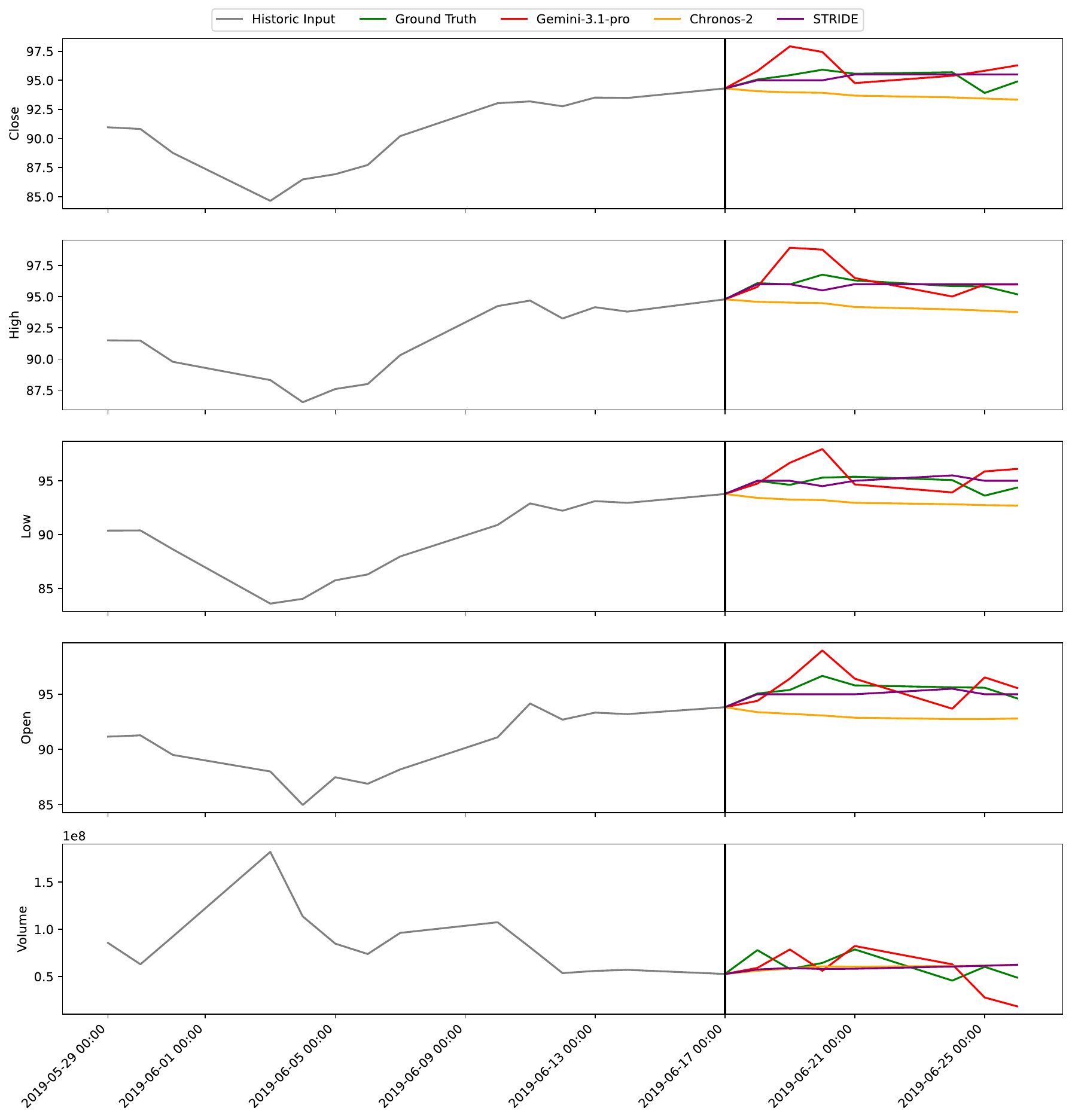} 
    \vspace{2mm} \\
    % \textit{Figure: Amazon Pricing forecasting comparison between Baseline (Gemini-3.1-pro) and Ours.}
\end{center}
\vspace{2mm}
\hrule
\vspace{3mm}
% --------------------------

\textbf{Baseline Reasoning (Gemini-3.1-pro)} \\
To forecast the next 7 steps for the 5 channels, we analyze the provided decomposition features. First, the Seasonality Component exhibits a strict 7-step repeating pattern. The next 7 steps will exactly mirror the seasonal values starting from the beginning of the cycle (e.g., the values from 2019-05-29 to 2019-06-06). Second, we evaluate the Trend Component. The price channels (Close, High, Low, Open) show a steady upward trajectory. By comparing the last few actual values to their seasonal adjustments, we estimate a conservative trend increment of approximately +0.30 per day for the price channels, starting from a base trend of roughly 95.00 for Close, 95.50 for High, 94.50 for Low, and 94.80 for Open. For Volume, the trend has been decreasing but appears to stabilize around 55,000,000 in the most recent actuals, so we project a flat baseline trend of 55,000,000. Finally, we synthesize the forecast by adding the projected linear trend to the exact repeating seasonal components for each respective step and channel.

\vspace{3mm}
\hrule
\vspace{3mm}

\textbf{\ours{}'s Reasoning}
\begin{itemize}
    \setlength{\itemsep}{1pt}
    \item \textbf{Cross-Channel Analysis:} The dataset comprises highly collinear daily price metrics (Open, High, Low, Close) and trading Volume for Amazon stock. Volume serves as a key indicator of market participation and momentum. The forecast indicates a tight daily trading range (spreads of roughly 0.5 to 1.0 points between High and Low) coupled with below-average volume (57M-62M shares compared to the historical mean of 85M). This dynamic suggests a low-volatility consolidation phase where the stock establishes a new, slightly higher baseline without aggressive institutional buying or panic selling pressure.
    \item \textbf{Trend Component Plan:} The price channels will exhibit a mild bullish drift, breaking just above the historical maximums (historically ~94.78) to establish a new support floor around 95.0. Open and Close prices will stabilize tightly in the 95.0 to 95.5 range, while the High tests near-term resistance at 96.0. The Low will steadily track between 94.5 and 95.5. Concurrently, Volume will demonstrate a gradual upward slope, climbing from approximately 57.4M to 62.4M shares over the forecast horizon, reflecting a slow, steady accumulation of market confidence at these newly established price levels.
    \item \textbf{Seasonality Component Plan:} While daily equity data is primarily driven by broader macroeconomic trends rather than strict daily seasonality, standard intra-week trading patterns apply. The forecast captures a typical trading week's rhythm, with prices remaining relatively flat and stable day-over-day. Volume will follow a standard structural trajectory, starting lower early in the period and gradually building momentum toward the end of the 7-day horizon, mimicking typical late-week portfolio adjustments and institutional positioning.
    \item \textbf{Residuals \& Event Plan:} Residuals will capture standard intraday market noise, bid-ask spread fluctuations, and minor algorithmic trading activity. The forecast assumes a quiet macroeconomic environment with no major catalyst events (such as earnings reports, product launches, or Federal Reserve announcements), which is explicitly reflected in the absence of volume spikes (remaining well below the historical 181M maximum). Minor deviations, such as the slight dip in the Low price to 94.5 on the third day, represent routine localized profit-taking or minor intraday sell-offs that are quickly absorbed by the prevailing baseline trend.
\end{itemize}

\vspace{3mm}
\hrule
\vspace{3mm}

\textbf{Evaluation Scores} \\
\textbf{Overall Scores:} \ours{}: 4.0 \quad|\quad Gemini-3.1-pro: 1.75

\vspace{2mm}

\textbf{Metrics Breakdown (Gemini-3.1-pro):}
\begin{itemize}
    \setlength{\itemsep}{1pt}
    \item \textbf{Domain Relevance (2/5):} The candidate uses generic time-series terminology (e.g., `Seasonality Component', `Trend Component', `linear trend') but completely lacks domain-specific financial language such as `market rally', `volatility', or 'liquidity' that would be appropriate for an equity dataset.
    \item \textbf{Forecasting Correctness (2/5):} The candidate incorrectly assumes a 'strict 7-step repeating pattern' for seasonality, ignoring the standard 5-day trading week for equities. It also completely misses the event-driven rally dynamics described in the Ground Truth, resulting in highly inaccurate predictions.
    \item \textbf{Event Relevance (2/5):} The candidate completely ignores the critical external events (the FOMC meeting and the San Francisco e-cigarette ban) that are the primary drivers of the market movement in the Ground Truth, relying instead on a flawed mechanical decomposition.
    \item \textbf{Logic Consistency (1/5):} There is a direct contradiction between the reasoning and the numbers. The candidate claims to mirror the seasonal values from 2019-05-29 to 2019-06-06 (a period where the actual Close price dropped sharply from ~90 to ~84) and add a small +0.30 daily trend. However, the predicted numerical values spike upwards to nearly 98, which mathematically contradicts the stated logic.
\end{itemize}

\vspace{2mm}

\textbf{Metrics Breakdown (\ours{}):}
\begin{itemize}
    \setlength{\itemsep}{1pt}
    \item \textbf{Domain Relevance (5/5):} The candidate demonstrates expert-level domain knowledge, using precise financial terminology such as `bullish breakout', `support floor', 'algorithmic selling', 'liquidity vacuums', and 'options expiration activity' to explain the market dynamics.
    \item \textbf{Forecasting Correctness (4/5):} The candidate captures the general price level (around 95-96) but completely misses the actual trend and volatility. It predicts a flat, low-volatility environment with a smooth volume curve, failing to capture the significant price peak on day 3, the sharp drop on day 6, and the erratic, high-volume surges present in the Ground Truth.
    \item \textbf{Event Relevance (2/5):} The candidate fails to identify the key macroeconomic event (the FOMC meeting) driving the market rally. Instead, it relies on generic, hypothetical events like 'minor positive catalyst', 'analyst upgrade', or 'options expiration' to justify its predicted movements, which are not grounded in the specific context of the Ground Truth.
    \item \textbf{Logic Consistency (5/5):} The numerical forecast perfectly aligns with the candidate's narrative. Every specific claim made in the text, such as the Close stepping up to 95.5, the High dipping to 95.5 mid-period, the Low dipping to 94.5 on day 3, and Volume dipping to 57.9M before rising to 62M, is exactly reflected in the generated numbers.
\end{itemize}

\end{tcolorbox}

%%%%%%%%%%%%%%%%%%%%%%%%%%%%%%%%%%%%%%%%%%%%%%%%%%%%%
% \clearpage

% \clearpage

\section{Prompts for the LLMs}
\label{app:prompts_llm}
In this section, we provide the prompts for the LLMs we used for \ours{} for reproducibility. For training data generation, we utilized the prompts from TFRBench~\cite{ahamed2026tfrbench}.

\begin{tcolorbox}[
    boxrule=0pt, 
    colframe=white, 
    colback=gray!5, 
    colbacktitle=black!50, 
    coltitle=black, 
    title={Prompt: TFRBench Search Agent  for Training Reasoning Generation (Event Retrieval)}, 
    breakable, 
    sharp corners
]

You are a specialized Search Agent. Your sole mission is to find relevant, time-specific external events that could impact a time-series forecast. Your task is to find events in BOTH the recent past and the future window.

\textbf{Core Search Parameters}
\begin{itemize}
    \item \textbf{Dataset Name:} \{dataset\_name\}
    \item \textbf{Recent Past Window:} \{past\_start\} to \{past\_end\}
    \item \textbf{Future Window:} \{future\_start\} to \{future\_end\}
    \item \textbf{Location Hint:} \{location\_hint\}
\end{itemize}

\textbf{Search Strategy}
\begin{enumerate}
    \item \textbf{Time-Bound:} Focus only on events within the specified windows.
    \item \textbf{Impactful Topics:} Search for objective, external factors like public holidays, major weather events (heatwaves, storms), sporting events, conferences, or economic announcements.
    \item \textbf{No Impact Analysis:} Do NOT analyze the `potential impact'. Just state the event and its date.
    \item \textbf{Prioritize:} Return only the top \{search\_events\} most significant, time-specific events. Do not return a long, unprioritized list.
\end{enumerate}

\textbf{IMPORTANT:} Your entire response must be a single, concise, numbered list of events.
\begin{itemize}
    \item Do NOT add any preamble (e.g., `Here are the events:').
    \item Format: 1. [Event Name] ([Date or Date Range])
    \item If no events are found, return the single word: None
\end{itemize}

Provide your findings directly.

\end{tcolorbox}

\begin{tcolorbox}[
    boxrule=0pt, 
    colframe=white, 
    colback=gray!5, 
    colbacktitle=black!50, 
    coltitle=black, 
    title={Prompt: TFRBench Reference Reasoning Generation}, 
    breakable, 
    sharp corners
]

You are a Master Forecaster. Your task is to generate a concrete, step-by-step reasoning plan for a downstream forecasting AI. You must analyze the historical data and the provided external events to create a strategic guide for predicting the future.

\textbf{Guidelines}
\begin{enumerate}
    \item \textbf{Direct Strategy:} Do not list assumptions. Go straight to the analysis.
    \item \textbf{Cross-Channel Analysis:} Explain how channels influence each other (e.g., `A rise in Channel A will likely drive a delayed rise in Channel B').
    \item \textbf{Qualitative \& Flexible:}
    \begin{itemize}
        \item Do NOT provide exact equations (e.g., `add 0.2317' is forbidden).
        \item Describe behavior (e.g., `The trend should continue its recent upward path, but at a decelerating rate').
        \item \textbf{CRITICAL:} You MUST include flexible numeric ranges/directions to make the reasoning concrete (e.g., ``expect an upward shift of roughly 10-15 units,'' ``volatility will likely double'').
    \end{itemize}
    \item \textbf{Event Integration:} You MUST integrate the provided `External Events'.
    \item \textbf{No Calculations:} Do not generate a mock forecast table.
\end{enumerate}

\textbf{--- Full Data Context ---}

\{full\_data\_context\_string\}

\textbf{--- Validated External Events ---}

\{external\_events\}

\textbf{IMPORTANT:} Your entire response must start \textit{immediately} with \texttt{Final Reasoning:}
\begin{itemize}
    \item Do NOT add any preamble.
    \item Follow this exact format:
\end{itemize}

Final Reasoning:\\
{} [Channel 1 Name]:
\begin{itemize}
    \item \textbf{Cross-Channel Analysis:} [Your analysis of dependencies.]
    \item \textbf{Trend Component Plan:} [Qualitative plan for the trend, including flexible numeric guidance.]
    \item \textbf{Seasonality Component Plan:} [Qualitative plan for seasonality.]
    \item \textbf{Residuals \& Event Plan:} [Qualitative plan for residuals and specific event impacts.]
\end{itemize}

[Channel 2 Name]:\\
... (Repeat for all channels)

\end{tcolorbox}

\begin{tcolorbox}[
    boxrule=0pt, 
    colframe=white, 
    colback=gray!5, 
    colbacktitle=black!50, 
    coltitle=black, 
    title={Prompt: Baseline Reasoning Provider (Large LLM)}, 
    breakable, 
    sharp corners
]

You are a Master Forecaster. Your task is to generate a concrete, step-by-step reasoning plan for a downstream forecasting AI. You must analyze the historical data context to create a strategic guide for predicting the future.

\textbf{Guidelines}
\begin{enumerate}
    \item \textbf{Direct Strategy:} Do not list assumptions. Go straight to the analysis.
    \item \textbf{Cross-Channel Analysis:} Explain how channels influence each other (e.g., `A rise in Channel A will likely drive a delayed rise in Channel B').
    \item \textbf{Qualitative \& Flexible:}
    \begin{itemize}
        \item Do NOT provide exact equations (e.g., `add 0.2317' is forbidden).
        \item Describe behavior (e.g., `The trend should continue its recent upward path, but at a decelerating rate').
        \item \textbf{CRITICAL:} You MUST include flexible numeric ranges/directions to make the reasoning concrete (e.g., ``expect an upward shift of roughly 10-15 units,'' ``volatility will likely double'').
    \end{itemize}
    \item \textbf{Event Prediction:} Since no external search is provided, you must infer and predict likely cyclical, seasonal, or external events purely based on the historical patterns. Describe their expected directional impact.
    \item \textbf{No Calculations:} Do not generate a mock forecast table.
\end{enumerate}

\textbf{--- Full Historical Data Context ---}

\{context\_str\}

\textbf{IMPORTANT:} Your entire response must start \textit{immediately} with \texttt{Final Reasoning:}
\begin{itemize}
    \item Do NOT add any preamble.
    \item Follow this exact format:
\end{itemize}

Final Reasoning:\\
{} [Channel 1 Name]:
\begin{itemize}
    \item \textbf{Cross-Channel Analysis:} [Your analysis of dependencies.]
    \item \textbf{Trend Component Plan:} [Qualitative plan for the trend, including flexible numeric guidance.]
    \item \textbf{Seasonality Component Plan:} [Qualitative plan for seasonality.]
    \item \textbf{Residuals \& Predicted Event Plan:} [Qualitative plan for residuals and inferred event impacts.]
\end{itemize}

[Channel 2 Name]:\\
... (Repeat for all channels)

\end{tcolorbox}

\begin{tcolorbox}[
    boxrule=0pt, 
    colframe=white, 
    colback=gray!5, % Light gray background for the text area (optional, set to white if preferred)
    colbacktitle=black!50, % Dark gray background for the title
    coltitle=black, % Black text for the title
    title={Prompt: Reasoning Generation (Small-LLM)}, 
    breakable, 
    sharp corners
]

You are a Master Forecaster. Your task is to generate a concrete, step-by-step reasoning plan for a downstream forecasting AI. You must analyze the historical data context to create a strategic guide for predicting the future.

\textbf{Guidelines}
\begin{enumerate}
    \item \textbf{Direct Strategy:} Do not list assumptions. Go straight to the analysis.
    \item \textbf{Cross-Channel Analysis:} Explain how channels influence each other.
    \item \textbf{Qualitative \& Flexible:} You MUST include flexible numeric ranges/directions (e.g., ``expect an upward shift of roughly 10-15 units''). Do NOT provide exact equations.
    \item \textbf{Event Prediction:} Infer likely cyclical, seasonal, or external events based purely on historical patterns.
    \item \textbf{Purely Analytical:} Do NOT invent physical causes, real-world events, or dates (e.g., do not guess it is ``cloud cover'' or ``equipment failure''). Your insight and planning must be derived strictly from the statistical properties of the numeric arrays provided.
    \item \textbf{No Code or Calculations:} Do not generate a mock forecast table. Do NOT write Python scripts or markdown code blocks.
    \item \textbf{Strict Formatting:} You must format your response exactly like this for every channel:
\end{enumerate}

[Channel X Name]:
\begin{itemize}
    \item \textbf{Cross-Channel Analysis:} [Your analysis]
    \item \textbf{Trend Component Plan:} [Your qualitative plan]
    \item \textbf{Seasonality Component Plan:} [Your qualitative plan]
    \item \textbf{Residuals \& Predicted Event Plan:} [Your qualitative plan]
\end{itemize}

\textbf{--- Full Historical Data Context ---}

\{context\_str\}

\textbf{--- Baseline Strategy ---}

\{baseline\_reasoning\}

\textbf{[Task]}

Refine the baseline strategy into the comprehensive reasoning plan based on the guidelines above. Encode the representations purely as plain text.

\end{tcolorbox}

\begin{tcolorbox}[
    boxrule=0pt, 
    colframe=white, 
    colback=gray!5, 
    colbacktitle=black!50, 
    coltitle=black, 
    title={Prompt: Baseline Reasoning Generation~\citep{ahamed2026tfrbench}}, 
    breakable, 
    sharp corners
]

Your task is to do time series forecasting WITH reasoning.

Given the input Data with context:

\{context\_str\}

Forecast the next \{pred\_len\} steps for the \{num\_channels\} channels.

\textbf{Strict Operational Protocols:}

1. Forecast a plausible continuation based on the signal structure.

2. \textbf{Scale Check}: Read the `Data Scale Reference' above. Your forecast must match this order of magnitude.

3. You are permitted to use the provided features, and you must reason over them.

4. Your output must be strictly limited to the final predicted values.

5. You must output step-by-step thinking, which is your reasoning. Your reasoning must be about pre-analysis. That is it should capture why certain forecast should be made rather than post explanation of the forecast.

6. Your reasoning is like a directive to an LLM, which will be used to improve the forecasting performance of a downstream LLM. Hence, it must be detailed and specific.

7. Utilize your reasoning first, then derive the forecast.

8. Provide the result ONLY as a JSON object containing the reasoning and numerical forecast array. Do not include any additional text.

9. YOUR REASONING IS NOT ABOUT THE POST ANALYSIS RATHER IT IS A FUTURE DIRECTION FOR THE DOWNSTREAM LLM TO FOLLOW.

Your entire response should consist of nothing but the JSON object.

\textbf{Required Output Specification:}

Your response must be a valid JSON object with exactly two keys:

- ``reasoning'': A detailed text string documenting your reasoning.

- ``forecast'': The 2D numerical array of shape (\{pred\_len\}, \{num\_channels\}) representing the final prediction.
\newline
\end{tcolorbox}

\begin{tcolorbox}[boxrule=0pt, 
    colframe=white, 
    colback=gray!5, 
    colbacktitle=black!50, 
    coltitle=black,  title=Prompt: Reasoning Evaluator (LLM-as-a-Judge), breakable, sharp corners]

You are an expert Time Series Forecasting Evaluator participating in the TFRBench Human Eval process. Your goal is to audit a Candidate Model's prediction by comparing its reasoning and numerical outputs against a Ground Truth analysis.

\textbf{1. Input Data}

\textbf{Context:} \{context\_str\}

\textbf{Ground Truth (The Ideal Analysis):}

Reasoning: \{ground\_truth\_reasoning\}

Actual Future Values: \{gt\_vals\_str\}

\textbf{Candidate Prediction (To Evaluate):}

Generated Reasoning: \{candidate\_reasoning\}

Predicted Values: \{cand\_vals\_str\}

\textbf{2. Task Annotation Instructions}

You must rate the Candidate Prediction on the following four metrics. Use the specific rubrics below to assign a score (1-5) for each.

\textbf{Metric 1: Domain Relevance (1-5)}
Does the reasoning incorporate domain-specific terminology and logic appropriate for the dataset context?
\begin{itemize}
    \item 1 (Irrelevant/Wrong): Wrong domain terminology. Logic makes no sense.
    \item 2 (Generic): Vague language without domain terms.
    \item 3 (Acceptable): Basic terms used correctly but lacks depth.
    \item 4 (Good): Specific terminology and standard forecasting logic used correctly.
    \item 5 (Expert): Deep domain expertise, precise jargon, matches Ground Truth logic.
\end{itemize}

\textbf{Metric 2: Forecasting Correctness (1-5)}
Does the reasoning correctly identify fundamental time-series dynamics (global trend, seasonality)?
\begin{itemize}
    \item 1 (Incorrect): Completely misses trend or seasonality.
    \item 2 (Weak): Identifies trend but ignores obvious seasonality.
    \item 3 (Average): Captures main trend/seasonality but misses magnitude/shape.
    \item 4 (Strong): Correctly identifies trend, seasonality, and general shape.
    \item 5 (Exact): Perfectly captures trend, seasonality, and inflection points.
\end{itemize}

\textbf{Metric 3: Event Relevance (1-5)}
Are external events factually grounded and causally relevant?
\begin{itemize}
    \item 1 (Hallucination): Mentions non-existent or false events.
    \item 2 (Irrelevant): Real events but no logical connection.
    \item 3 (Correlated): Identifies event but explanation is vague.
    \item 4 (Causal): Clearly links event to specific data movement.
    \item 5 (Aligned): Detailed, accurate causal explanation of impact.
\end{itemize}

\textbf{Metric 4: Logic-to-Number Consistency (1-5)}
Does the narrative explanation logically support the generated numerical forecast?
\begin{itemize}
    \item 1 (Contradiction): Text and numbers are opposites.
    \item 2 (Disconnected): Text describes shapes not seen in numbers.
    \item 3 (Weak Consistency): Direction matches but magnitude is off.
    \item 4 (Consistent): Numbers generally follow the narrative.
    \item 5 (Alignment): Numerical forecast is a precise translation of the reasoning.
\end{itemize}

\textbf{3. Output Format}

Provide your assessment as a single valid JSON object. Do not include any text before or after the JSON.

\{\\
    ``metric\_1\_domain\_relevance'': \{ ``score'': $<$int 1-5$>$, ``reasoning'': ``...'' \},\\
    ``metric\_2\_forecasting\_correctness'': \{ ``score'': $<$int 1-5$>$, ``reasoning'': ``...'' \},\\
    ``metric\_3\_event\_relevance'': \{ ``score'': $<$int 1-5$>$, ``reasoning'': ``...'' \},\\
    ``metric\_4\_logic\_consistency'': \{ ``score'': $<$int 1-5$>$, ``reasoning'': ``...'' \},\\
    ``final\_critique'': ``$<$Summary of strengths/weaknesses$>$''\\
\}

\end{tcolorbox}

\clearpage
\section{Forecasting Performance Evaluation on TFRBench}
\label{app:tfrbench_numerical}
Tables \ref{tab:numerical_tfrbench_indomain} and \ref{tab:numerical_tfrbench_ood} present the numerical evaluation on TFRBench across in-domain and out-of-domain settings. Overall, our reasoning-aware framework establishes state-of-the-art accuracy, achieving the lowest average MASE of 0.615 (in-domain) and 0.724 (out-of-domain). By injecting semantic reasoning as a structural prior, \ours{} generally reduces error margins compared to unguided foundation models like Chronos-2.0 and TimesFM-2.5. It particularly excels on datasets like Traffic and NYC Taxi, while entirely bypassing the mathematical degradation inherent in zero-shot LLM forecasters. 
However, we observe a distinct boundary condition in highly stochastic, event-driven domains like the Amazon and Apple datasets, where our framework marginally underperforms. This localized degradation is a direct artifact of our closed-loop inference architecture. Because during inference, the student LLM generates its reasoning prior based strictly on historical numerical context, lacking access to real-time external covariates like macroeconomic news, it defaults to predicting conservative, low-volatility continuations. Consequently, projecting this over-smoothed reasoning prior into the TSFM inadvertently dampens the numerical encoder's responsiveness to sudden market shocks. This reveals a necessity for dynamic external knowledge in financial forecasting.

\begin{table*}[h]
\centering
\caption{Numerical performance on TFRBench (In-Domain datasets). We report Mean Absolute Error (MAE) and Mean Absolute Scaled Error (MASE) in the format $mean_{std}$ (lower is better). To avoid cross-dataset scaling issues, we only present the average for MASE. \textbf{Bold} indicates the best performance, and \underline{underline} indicates the second best.}
\resizebox{\linewidth}{!}{%
\begin{tabular}{@{}l|r@{}lr@{}l|r@{}lr@{}l|r@{}lr@{}l|r@{}lr@{}l|r@{}lr@{}l|r@{}l@{}}
\toprule
\multirow{2}{*}{\textbf{Models}} & \multicolumn{4}{c|}{\textbf{Solar Daily}} & \multicolumn{4}{c|}{\textbf{Electricity}} & \multicolumn{4}{c|}{\textbf{Car Parts}} & \multicolumn{4}{c|}{\textbf{Hierarchical Sales}} & \multicolumn{4}{c|}{\textbf{Bitbrains Fast Storage}} & \multicolumn{2}{c}{\textbf{Average}} \\ \cmidrule(l){2-23}
 & \multicolumn{2}{c}{\textbf{MAE}} & \multicolumn{2}{c|}{\textbf{MASE}} & \multicolumn{2}{c}{\textbf{MAE}} & \multicolumn{2}{c|}{\textbf{MASE}} & \multicolumn{2}{c}{\textbf{MAE}} & \multicolumn{2}{c|}{\textbf{MASE}} & \multicolumn{2}{c}{\textbf{MAE}} & \multicolumn{2}{c|}{\textbf{MASE}} & \multicolumn{2}{c}{\textbf{MAE}} & \multicolumn{2}{c|}{\textbf{MASE}} & \multicolumn{2}{c}{\textbf{MASE}} \\ \midrule
ARIMA & $1.444$ & $_{0.000}$ & $0.699$ & $_{0.000}$ & $5.564$ & $_{0.000}$ & $1.536$ & $_{0.000}$ & $0.569$ & $_{0.000}$ & $0.873$ & $_{0.000}$ & $2.239$ & $_{0.000}$ & $0.789$ & $_{0.000}$ & $3.57 \times 10^{4}$ & $_{0.000}$ & $0.799$ & $_{0.000}$ & $0.939$ & $_{0.000}$ \\
TimesFM-2.5 & $1.475$ & $_{0.000}$ & $0.735$ & $_{0.000}$ & $4.758$ & $_{0.000}$ & $1.380$ & $_{0.000}$ & $0.328$ & $_{0.000}$ & $0.448$ & $_{0.000}$ & $2.142$ & $_{0.000}$ & $0.753$ & $_{0.000}$ & $\underline{2.62 \times 10^{4}}$ & $_{\underline{0.000}}$ & $3.401$ & $_{0.000}$ & $1.343$ & $_{0.000}$ \\
Chronos-2.0 & $\underline{1.409}$ & $_{\underline{0.000}}$ & $\underline{0.685}$ & $_{\underline{0.000}}$ & $4.162$ & $_{0.000}$ & $1.271$ & $_{0.000}$ & $\underline{0.294}$ & $_{\underline{0.000}}$ & $\underline{0.413}$ & $_{\underline{0.000}}$ & $\underline{2.127}$ & $_{\underline{0.000}}$ & $\underline{0.750}$ & $_{\underline{0.000}}$ & $2.69 \times 10^{4}$ & $_{0.000}$ & $\underline{0.708}$ & $_{\underline{0.000}}$ & $\underline{0.765}$ & $_{\underline{0.000}}$ \\
\midrule
Gemini-2.5-Flash & $2.166$ & $_{0.432}$ & $1.037$ & $_{0.151}$ & $6.271$ & $_{0.315}$ & $1.785$ & $_{0.115}$ & $0.460$ & $_{0.081}$ & $0.662$ & $_{0.105}$ & $2.675$ & $_{0.026}$ & $0.904$ & $_{0.035}$ & $4.62 \times 10^{4}$ & $_{636.396}$ & $1.110$ & $_{0.096}$ & $1.100$ & $_{0.101}$ \\
Gemini-2.5-Pro & $2.598$ & $_{0.288}$ & $1.217$ & $_{0.123}$ & $7.604$ & $_{2.577}$ & $2.244$ & $_{0.862}$ & $0.432$ & $_{0.119}$ & $0.630$ & $_{0.192}$ & $2.775$ & $_{0.068}$ & $0.938$ & $_{0.029}$ & $4.99 \times 10^{4}$ & $_{5.80 \times 10^{3}}$ & $1.266$ & $_{0.097}$ & $1.259$ & $_{0.261}$ \\
Claude-Sonnet-4 & $2.481$ & $_{0.475}$ & $1.169$ & $_{0.242}$ & $8.774$ & $_{0.477}$ & $2.647$ & $_{0.014}$ & $0.475$ & $_{0.111}$ & $0.703$ & $_{0.187}$ & $3.417$ & $_{0.899}$ & $1.125$ & $_{0.293}$ & $1.11 \times 10^{5}$ & $_{8.44 \times 10^{4}}$ & $3.026$ & $_{2.331}$ & $1.734$ & $_{0.613}$ \\
Claude-Sonnet-4.5 & $1.780$ & $_{0.209}$ & $0.869$ & $_{0.078}$ & $6.316$ & $_{0.511}$ & $1.714$ & $_{0.202}$ & $0.413$ & $_{0.021}$ & $0.595$ & $_{0.022}$ & $2.710$ & $_{0.289}$ & $0.898$ & $_{0.107}$ & $8.34 \times 10^{4}$ & $_{6.02 \times 10^{4}}$ & $3.244$ & $_{3.262}$ & $1.464$ & $_{0.734}$ \\
Gemini-3.1-Pro & $1.603$ & $_{0.011}$ & $0.779$ & $_{0.010}$ & $\underline{3.944}$ & $_{\underline{0.076}}$ & $\underline{1.184}$ & $_{\underline{0.115}}$ & $0.476$ & $_{0.020}$ & $0.671$ & $_{0.052}$ & $2.341$ & $_{0.033}$ & $0.801$ & $_{0.022}$ & $4.16 \times 10^{4}$ & $_{353.553}$ & $0.833$ & $_{0.004}$ & $0.854$ & $_{0.040}$ \\
\ours{} (+Chronos-2.0)& $\mathbf{0.884}$ & $_{\mathbf{0.051}}$ & $\mathbf{0.448}$ & $_{\mathbf{0.019}}$ & $\mathbf{2.922}$ & $_{\mathbf{0.192}}$ & $\mathbf{0.828}$ & $_{\mathbf{0.015}}$ & $\mathbf{0.280}$ & $_{\mathbf{0.002}}$ & $\mathbf{0.404}$ & $_{\mathbf{0.003}}$ & $\mathbf{2.081}$ & $_{\mathbf{0.075}}$ & $\mathbf{0.719}$ & $_{\mathbf{0.008}}$ & $\mathbf{2.47 \times 10^{4}}$ & $_{\mathbf{1.09 \times 10^{3}}}$ & $\mathbf{0.676}$ & $_{\mathbf{0.006}}$ & $\mathbf{0.615}$ & $_{\mathbf{0.010}}$ \\
\bottomrule
\end{tabular}
}
\label{tab:numerical_tfrbench_indomain}
\end{table*}

\begin{table*}[h]
\centering
\caption{Numerical performance on TFRBench (Out-of-Domain datasets). We report Mean Absolute Error (MAE) and Mean Absolute Scaled Error (MASE) in the format $mean_{std}$ (lower is better). To avoid cross-dataset scaling issues, we only present the average for MASE. \textbf{Bold} indicates the best performance, and \underline{underline} indicates the second best.}
\resizebox{\textwidth}{!}{%
\begin{tabular}{@{}l|r@{}lr@{}l|r@{}lr@{}l|r@{}lr@{}l|r@{}lr@{}l|r@{}lr@{}l|r@{}l@{}}
\toprule
\multirow{2}{*}{\textbf{Models}} & \multicolumn{4}{c|}{\textbf{Web Traffic}} & \multicolumn{4}{c|}{\textbf{Traffic}} & \multicolumn{4}{c|}{\textbf{Nyc Taxi}} & \multicolumn{4}{c|}{\textbf{Amazon}} & \multicolumn{4}{c|}{\textbf{Apple}} & \multicolumn{2}{c}{\textbf{Average}} \\ \cmidrule(l){2-23}
 & \multicolumn{2}{c}{\textbf{MAE}} & \multicolumn{2}{c|}{\textbf{MASE}} & \multicolumn{2}{c}{\textbf{MAE}} & \multicolumn{2}{c|}{\textbf{MASE}} & \multicolumn{2}{c}{\textbf{MAE}} & \multicolumn{2}{c|}{\textbf{MASE}} & \multicolumn{2}{c}{\textbf{MAE}} & \multicolumn{2}{c|}{\textbf{MASE}} & \multicolumn{2}{c}{\textbf{MAE}} & \multicolumn{2}{c|}{\textbf{MASE}} & \multicolumn{2}{c}{\textbf{MASE}} \\ \midrule
ARIMA & $15.400$ & $_{0.000}$ & $0.788$ & $_{0.000}$ & $0.039$ & $_{0.000}$ & $2.628$ & $_{0.000}$ & $8.65 \times 10^{3}$ & $_{0.000}$ & $3.429$ & $_{0.000}$ & $9.11 \times 10^{6}$ & $_{0.000}$ & $0.995$ & $_{0.000}$ & $2.28 \times 10^{7}$ & $_{0.000}$ & $1.100$ & $_{0.000}$ & $1.788$ & $_{0.000}$ \\
TimesFM-2.5 & $\underline{12.000}$ & $_{\underline{0.000}}$ & $\underline{0.676}$ & $_{\underline{0.000}}$ & $0.014$ & $_{0.000}$ & $0.847$ & $_{0.000}$ & $2.70 \times 10^{3}$ & $_{0.000}$ & $1.051$ & $_{0.000}$ & $\mathbf{6.56 \times 10^{6}}$ & $_{\mathbf{0.000}}$ & $\mathbf{0.747}$ & $_{\mathbf{0.000}}$ & $\mathbf{1.86 \times 10^{7}}$ & $_{\mathbf{0.000}}$ & $\mathbf{0.879}$ & $_{\mathbf{0.000}}$ & $0.840$ & $_{0.000}$ \\
Chronos-2.0 & $\mathbf{11.900}$ & $_{\mathbf{0.000}}$ & $0.679$ & $_{0.000}$ & $\underline{0.011}$ & $_{\underline{0.000}}$ & $\underline{0.664}$ & $_{\underline{0.000}}$ & $1.96 \times 10^{3}$ & $_{0.000}$ & $\underline{0.790}$ & $_{\underline{0.000}}$ & $7.24 \times 10^{6}$ & $_{0.000}$ & $0.821$ & $_{0.000}$ & $\underline{1.95 \times 10^{7}}$ & $_{\underline{0.000}}$ & $\underline{0.936}$ & $_{\underline{0.000}}$ & $\underline{0.778}$ & $_{\underline{0.000}}$ \\
\midrule
Gemini-2.5-Flash & $23.300$ & $_{0.000}$ & $1.120$ & $_{0.000}$ & $0.027$ & $_{0.000}$ & $1.647$ & $_{0.000}$ & $5.75 \times 10^{3}$ & $_{0.000}$ & $2.289$ & $_{0.000}$ & $1.37 \times 10^{7}$ & $_{0.000}$ & $1.333$ & $_{0.001}$ & $3.37 \times 10^{7}$ & $_{0.000}$ & $1.482$ & $_{0.000}$ & $1.574$ & $_{0.000}$ \\
Gemini-2.5-Pro & $28.200$ & $_{1.000}$ & $1.538$ & $_{0.046}$ & $0.030$ & $_{0.001}$ & $1.885$ & $_{0.025}$ & $7.10 \times 10^{3}$ & $_{110.000}$ & $2.815$ & $_{0.033}$ & $1.49 \times 10^{7}$ & $_{3.00 \times 10^{5}}$ & $1.464$ & $_{0.015}$ & $3.57 \times 10^{7}$ & $_{6.00 \times 10^{5}}$ & $1.589$ & $_{0.036}$ & $1.858$ & $_{0.031}$ \\
Claude-Sonnet-4 & $19.900$ & $_{0.400}$ & $1.211$ & $_{0.040}$ & $0.024$ & $_{0.004}$ & $1.450$ & $_{0.168}$ & $5.31 \times 10^{3}$ & $_{120.000}$ & $2.085$ & $_{0.055}$ & $1.34 \times 10^{7}$ & $_{2.00 \times 10^{5}}$ & $1.511$ & $_{0.050}$ & $3.54 \times 10^{7}$ & $_{9.00 \times 10^{5}}$ & $1.651$ & $_{0.031}$ & $1.582$ & $_{0.069}$ \\
Claude-Sonnet-4.5 & $18.000$ & $_{0.800}$ & $0.865$ & $_{0.003}$ & $0.024$ & $_{0.000}$ & $1.458$ & $_{0.001}$ & $4.63 \times 10^{3}$ & $_{40.000}$ & $1.837$ & $_{0.015}$ & $7.97 \times 10^{6}$ & $_{1.20 \times 10^{5}}$ & $0.870$ & $_{0.011}$ & $2.16 \times 10^{7}$ & $_{2.00 \times 10^{5}}$ & $0.989$ & $_{0.011}$ & $1.204$ & $_{0.008}$ \\
Gemini-3.1-Pro & $13.600$ & $_{0.000}$ & $1.472$ & $_{0.000}$ & $0.015$ & $_{0.000}$ & $1.102$ & $_{0.000}$ & $\underline{1.73 \times 10^{3}}$ & $_{\underline{0.000}}$ & $0.889$ & $_{0.000}$ & $1.03 \times 10^{7}$ & $_{0.000}$ & $1.776$ & $_{0.000}$ & $2.70 \times 10^{7}$ & $_{0.000}$ & $1.802$ & $_{0.000}$ & $1.408$ & $_{0.000}$ \\
\ours{} (+Chronos-2.0)& $12.076$ & $_{0.175}$ & $\mathbf{0.664}$ & $_{\mathbf{0.001}}$ & $\mathbf{0.011}$ & $_{\mathbf{0.000}}$ & $\mathbf{0.634}$ & $_{\mathbf{0.002}}$ & $\mathbf{1.39 \times 10^{3}}$ & $_{\mathbf{74.878}}$ & $\mathbf{0.555}$ & $_{\mathbf{0.027}}$ & $\underline{6.56 \times 10^{6}}$ & $_{\underline{1.98 \times 10^{4}}}$ & $\underline{0.791}$ & $_{\underline{0.003}}$ & $1.99 \times 10^{7}$ & $_{3.62 \times 10^{5}}$ & $0.973$ & $_{0.020}$ & $\mathbf{0.724}$ & $_{\mathbf{0.011}}$ \\
\bottomrule
\end{tabular}
}
\label{tab:numerical_tfrbench_ood}
\end{table*}

\clearpage
\section{Reasoning Performance Evaluation on TFRBench}
\label{app:tfrbench_all_reason}
% Requires \usepackage{booktabs}, \usepackage{multirow}, and \usepackage{graphicx}
\begin{table}[h]
\centering
\footnotesize
\caption{\textbf{LLM-as-a-Judge Results (In-Domain).} We report the reasoning quality scores comparing against the baseline models. Scores are measured on a scale of 1 to 5, where higher values indicate better performance. The metrics cover four dimensions: Domain Relevance (\textbf{Dom.}), Forecasting Correctness (\textbf{Fcst.}), Event Relevance (\textbf{Evt.}), and Logic-to-Number Consistency (\textbf{Logic}).}
\resizebox{0.7\linewidth}{!}{%
\begin{tabular}{@{}c|l|cccc@{}}
\toprule
\multicolumn{1}{c|}{\multirow{2}{*}{\textbf{Dataset}}} & \multicolumn{1}{c|}{\multirow{2}{*}{\textbf{Models}}} & \multicolumn{4}{c}{\textbf{Metrics}} \\ \cmidrule(l){3-6}
\multicolumn{1}{c|}{} & \multicolumn{1}{c|}{} & \textbf{Dom.} & \textbf{Fcst.} & \textbf{Evt.} & \textbf{Logic} \\ \midrule

\multirow{7}{*}{Solar Daily}
 & Gemini-2.5-Flash & $2.472$ & $2.399$ & $1.933$ & $4.435$ \\
 & Gemini-2.5-Pro & $3.026$ & $2.316$ & $2.036$ & $4.528$ \\
 & Claude-Sonnet-4 & $4.503$ & $2.570$ & $2.503$ & $2.870$ \\
 & Claude-Sonnet-4.5 & $4.922$ & $3.062$ & $2.891$ & $3.363$ \\
 & Gemini-3.1-Pro & $3.736$ & $2.995$ & $2.109$ & $4.974$ \\
 & Gemma-3-4B-it & $2.731$ & $2.642$ & $2.047$ & $1.119$ \\
 & \ours{} (+Chronos-2.0) & $4.896$ & $3.093$ & $2.062$ & $4.839$ \\
\midrule

\multirow{7}{*}{Electricity}
 & Gemini-2.5-Flash & $2.856$ & $2.229$ & $1.136$ & $4.643$ \\
 & Gemini-2.5-Pro & $2.331$ & $2.739$ & $1.725$ & $4.569$ \\
 & Claude-Sonnet-4 & $3.765$ & $1.955$ & $1.439$ & $2.977$ \\
 & Claude-Sonnet-4.5 & $3.810$ & $2.054$ & $2.093$ & $3.167$ \\
 & Gemini-3.1-Pro & $3.091$ & $3.006$ & $2.156$ & $4.972$ \\
 & Gemma-3-4B-it & $2.612$ & $2.459$ & $2.008$ & $1.278$ \\
 & \ours{} (+Chronos-2.0) & $4.416$ & $3.671$ & $2.193$ & $4.881$ \\
\midrule

\multirow{7}{*}{Car Parts}
 & Gemini-2.5-Flash & $2.244$ & $2.383$ & $1.764$ & $4.845$ \\
 & Gemini-2.5-Pro & $3.123$ & $2.658$ & $1.863$ & $4.181$ \\
 & Claude-Sonnet-4 & $3.760$ & $2.605$ & $1.996$ & $3.053$ \\
 & Claude-Sonnet-4.5 & $4.120$ & $2.675$ & $2.022$ & $3.384$ \\
 & Gemini-3.1-Pro & $3.506$ & $2.341$ & $1.808$ & $4.949$ \\
 & Gemma-3-4B-it & $2.380$ & $1.958$ & $1.874$ & $1.283$ \\
 & \ours{} (+Chronos-2.0) & $4.783$ & $3.503$ & $2.374$ & $4.993$ \\
\midrule

\multirow{7}{*}{Hierarchical Sales}
 & Gemini-2.5-Flash & $2.666$ & $2.151$ & $1.035$ & $4.753$ \\
 & Gemini-2.5-Pro & $2.619$ & $2.974$ & $1.622$ & $4.725$ \\
 & Claude-Sonnet-4 & $3.091$ & $1.974$ & $1.153$ & $3.318$ \\
 & Claude-Sonnet-4.5 & $3.757$ & $2.613$ & $2.046$ & $3.651$ \\
 & Gemini-3.1-Pro & $2.887$ & $2.916$ & $1.977$ & $4.636$ \\
 & Gemma-3-4B-it & $2.309$ & $2.419$ & $1.869$ & $1.041$ \\
 & \ours{} (+Chronos-2.0) & $4.127$ & $2.991$ & $2.059$ & $4.950$ \\
\midrule

\multirow{7}{*}{Bitbrains Fast Storage}
 & Gemini-2.5-Flash & $3.174$ & $2.606$ & $2.699$ & $3.152$ \\
 & Gemini-2.5-Pro & $3.807$ & $2.565$ & $2.944$ & $3.845$ \\
 & Claude-Sonnet-4 & $3.942$ & $2.240$ & $3.122$ & $1.988$ \\
 & Claude-Sonnet-4.5 & $4.714$ & $2.780$ & $3.428$ & $2.507$ \\
 & Gemini-3.1-Pro & $3.555$ & $3.112$ & $2.926$ & $4.985$ \\
 & Gemma-3-4B-it & $2.233$ & $2.028$ & $2.307$ & $1.322$ \\
 & \ours{} (+Chronos-2.0) & $4.897$ & $3.221$ & $2.959$ & $4.913$ \\
\bottomrule
\end{tabular}
}
\label{tab:judge-performance-indomain}
\end{table}
\begin{table}
\centering
\footnotesize
\caption{\textbf{LLM-as-a-Judge Results (Out-of-Domain).} We report the reasoning quality scores comparing against the baseline models. Scores are measured on a scale of 1 to 5, where higher values indicate better performance. The metrics cover four dimensions: Domain Relevance (\textbf{Dom.}), Forecasting Correctness (\textbf{Fcst.}), Event Relevance (\textbf{Evt.}), and Logic-to-Number Consistency (\textbf{Logic}).}
\resizebox{0.7\linewidth}{!}{%
\begin{tabular}{@{}c|l|cccc@{}}
\toprule
\multicolumn{1}{c|}{\multirow{2}{*}{\textbf{Dataset}}} & \multicolumn{1}{c|}{\multirow{2}{*}{\textbf{Models}}} & \multicolumn{4}{c}{\textbf{Metrics}} \\ \cmidrule(l){3-6}
\multicolumn{1}{c|}{} & \multicolumn{1}{c|}{} & \textbf{Dom.} & \textbf{Fcst.} & \textbf{Evt.} & \textbf{Logic} \\ \midrule

\multirow{7}{*}{Web Traffic}
 & Gemini-2.5-Flash & $3.158$ & $2.904$ & $1.961$ & $4.403$ \\
 & Gemini-2.5-Pro & $3.585$ & $2.831$ & $2.039$ & $4.693$ \\
 & Claude-Sonnet-4 & $3.749$ & $2.104$ & $2.080$ & $3.100$ \\
 & Claude-Sonnet-4.5 & $3.928$ & $2.325$ & $2.120$ & $3.727$ \\
 & Gemini-3.1-Pro & $3.100$ & $3.180$ & $2.100$ & $4.990$ \\
 & Gemma-3-4B-it & $2.196$ & $2.501$ & $1.911$ & $1.044$ \\
 & \ours{} (+Chronos-2.0) & $4.045$ & $3.016$ & $2.085$ & $4.969$ \\
\midrule

\multirow{7}{*}{Traffic}
 & Gemini-2.5-Flash & $2.708$ & $1.365$ & $1.437$ & $4.416$ \\
 & Gemini-2.5-Pro & $2.762$ & $1.946$ & $1.896$ & $4.584$ \\
 & Claude-Sonnet-4 & $3.535$ & $2.010$ & $1.468$ & $2.946$ \\
 & Claude-Sonnet-4.5 & $4.294$ & $2.254$ & $2.488$ & $3.586$ \\
 & Gemini-3.1-Pro & $3.690$ & $2.740$ & $2.350$ & $5.000$ \\
 & Gemma-3-4B-it & $2.882$ & $2.307$ & $2.192$ & $1.670$ \\
 & \ours{} (+Chronos-2.0) & $4.726$ & $3.394$ & $2.619$ & $4.964$ \\
\midrule

\multirow{7}{*}{Nyc Taxi}
 & Gemini-2.5-Flash & $2.808$ & $1.248$ & $1.176$ & $4.327$ \\
 & Gemini-2.5-Pro & $2.814$ & $1.922$ & $1.703$ & $4.360$ \\
 & Claude-Sonnet-4 & $3.576$ & $2.078$ & $1.327$ & $2.868$ \\
 & Claude-Sonnet-4.5 & $4.532$ & $2.297$ & $2.363$ & $2.767$ \\
 & Gemini-3.1-Pro & $4.040$ & $3.380$ & $2.280$ & $4.910$ \\
 & Gemma-3-4B-it & $2.762$ & $2.018$ & $2.005$ & $1.141$ \\
 & \ours{} (+Chronos-2.0) & $4.811$ & $3.465$ & $2.215$ & $4.870$ \\
\midrule

\multirow{7}{*}{Amazon}
 & Gemini-2.5-Flash & $2.082$ & $1.473$ & $1.625$ & $3.624$ \\
 & Gemini-2.5-Pro & $2.262$ & $1.488$ & $1.637$ & $3.213$ \\
 & Claude-Sonnet-4 & $2.686$ & $1.680$ & $1.677$ & $2.811$ \\
 & Claude-Sonnet-4.5 & $3.588$ & $2.012$ & $1.893$ & $3.970$ \\
 & Gemini-3.1-Pro & $2.730$ & $1.640$ & $1.670$ & $3.840$ \\
 & Gemma-3-4B-it & $1.948$ & $1.290$ & $1.899$ & $1.146$ \\
 & \ours{} (+Chronos-2.0) & $4.494$ & $1.790$ & $1.906$ & $4.954$ \\
\midrule

\multirow{7}{*}{Apple}
 & Gemini-2.5-Flash & $2.069$ & $1.523$ & $1.613$ & $3.513$ \\
 & Gemini-2.5-Pro & $2.217$ & $1.557$ & $1.695$ & $3.380$ \\
 & Claude-Sonnet-4 & $2.525$ & $1.680$ & $1.601$ & $2.608$ \\
 & Claude-Sonnet-4.5 & $3.617$ & $2.075$ & $1.917$ & $3.786$ \\
 & Gemini-3.1-Pro & $2.670$ & $1.640$ & $1.640$ & $3.850$ \\
 & Gemma-3-4B-it & $1.925$ & $1.354$ & $1.898$ & $1.194$ \\
 & \ours{} (+Chronos-2.0) & $4.542$ & $1.621$ & $1.947$ & $4.914$ \\
\bottomrule
\end{tabular}
}
\label{tab:judge-performance-ood}
\end{table}

%%%%%%%%%%%%%%%%%%%%%%%%%%%%%%%%%%%%%%%%%%%%%%%%%%%%%%%%%%%%

\end{document}